\documentclass[conference]{IEEEtran}
\IEEEoverridecommandlockouts
\usepackage{cite}
\usepackage{amsmath,amssymb,amsfonts}
\usepackage{algorithmic}
\usepackage{graphicx}
\usepackage{textcomp}
\usepackage{dsfont}
\usepackage{xcolor}
\def\BibTeX{{\rm B\kern-.05em{\sc i\kern-.025em b}\kern-.08em
		T\kern-.1667em\lower.7ex\hbox{E}\kern-.125emX}}
\begin{document}
	
	\title{Measuring Overfitting in Convolutional Neural Networks using Adversarial Perturbations and Label Noise
	}
	
	\author{\IEEEauthorblockN{Svetlana Pavlitskaya$^{1}$, Joël Oswald$^{2}$, J.~Marius Zöllner$^{1,2}$}
		\IEEEauthorblockA{\textit{$^{1}$ FZI Research Center for Information Technology} \\
			\textit{$^{2}$ Karlsruhe Institute of Technology (KIT)}\\
			Karlsruhe, Germany \\
			pavlitskaya@fzi.de}
	}
	\maketitle
	
	\begin{abstract}
		Although numerous methods to reduce the overfitting of convolutional neural networks (CNNs) exist, it is still not clear how to confidently measure the degree of overfitting. A metric reflecting the overfitting level might be, however, extremely helpful for the comparison of different architectures and for the evaluation of various techniques to tackle overfitting. Motivated by the fact that overfitted neural networks tend to rather memorize noise in the training data than generalize to unseen data, we examine how the training accuracy changes in the presence of increasing data perturbations and study the connection to overfitting. While previous work focused on label noise only, we examine a spectrum of techniques to inject noise into the training data, including adversarial perturbations and input corruptions. Based on this, we define two new metrics that can confidently distinguish between correct and overfitted models. For the evaluation, we derive a pool of models for which the overfitting behavior is known beforehand. To test the effect of various factors, we introduce several anti-overfitting measures in architectures based on VGG and ResNet and study their impact, including regularization techniques, training set size, and the number of parameters. Finally, we assess the applicability of the proposed metrics by measuring the overfitting degree of several CNN architectures outside of our model pool.
	\end{abstract}
	
	
	\section{Introduction}
	Overfitting is a known danger for machine learning algorithms and convolutional neural networks (CNNs) in particular. An overfitted learner performs well on the training data but fails to perform as well as expected on new data. Not only is overfitting malicious because it leads to unexpectedly low performance on unseen data. A further menace was recently addressed by Recht et al.~\cite{recht2018cifar} -- due to repeated usage of the same test sets in popular datasets and benchmarks, the evaluation of model selection is questionable. Moreover, a close relationship between overfitting and privacy risks is discussed by Yeom et al.~\cite{yeom2018privacy} -- overfitted neural networks might leak sensitive data that were used to train the model. 
	
	Currently, no uniform approach to measure overfitting exists. The standard technique (difference between training and validation set accuracies) is evidently limited. First, the validation set might not be representative of the underlying data distribution, which would lead to a false estimation of the true accuracy. Second, this approach does not apply to comparing different models trained on various datasets.
	
	Recently, two novel approaches to measure overfitting have been proposed. First, the Perturbed Model Validation approach by Zhang et al.~\cite{zhang2019perturbed} consists in modifying training data by injecting random noise into labels and retraining models on this dataset. The authors show, that their method helps to separate correct models from overfitted ones -- the detailed evaluation on neural networks is, however, missing. Second, Werpachowski et al.~\cite{werpachowski2019detecting} claim that adversarial manipulation of inputs enables precise estimation of the true performance of a model. The authors focus, however, on overfitting to a test set. 
	
	We start by evaluating these two techniques on a pool of CNN models with different architectures and training hyperparameters. We then propose several further data modification methods and define several metrics based on those.
	
	Our contributions can be summarized as follows:
	\begin{itemize}
		\item We apply approaches from~\cite{werpachowski2019detecting} and~\cite{zhang2019perturbed} to CNNs to examine the overfitting to a training dataset.
		\item We extend and enhance these methods with further techniques to modify training data.
		\item We propose two metrics to measure the overfitting of CNNs and support the model selection process.
		\item We extensively evaluate all approaches and metrics on a pool of overall 16 CNN models with various anti-overfitting features.
	\end{itemize}
	
	\section{Background and Related Work}
	\subsection{Overfitting}
	
	In statistical machine learning, the learning problem consists in finding a hypothesis  $h$, which is a function $\mathcal{X} \to \mathcal{Y}$ for a domain set $\mathcal{X}$, where each instance $x \in \mathcal{X}$ (generated by some probability distribution $\mathcal{D}$) has a label $y$ of the label set $\mathcal{Y}$. $h$ can then be used to predict labels for new, unseen instances of $\mathcal{X}$.
	
	The true relationship $\mathnormal{f^*}: \mathcal{X} \to \mathcal{Y}$ between $\mathcal{X}$ and $\mathnormal{Y}$ is unknown,  the learner only has access to the training set $S$, which consists of tuples $(x_i, y_i),\text{ } i=1 \dots m$. We assume, that the instances of $S$ are identically distributed, according to $\mathcal{D}$, and labeled by $\mathnormal{f^*}$. The learner returns a hypothesis $h$ based on this training data solely. 
	
	Since the data distribution $\mathcal{D}$ and the true labels, as defined by $f^*$, are not known, it is not possible to calculate the true generalization error (or risk)  $R_{\mathcal{D}, f^*}(h)$ of the hypothesis $h$ (Equation \ref{eq:truerisk}) as well as its accuracy ${Acc}(\mathcal{D}, h)$ (Equation \ref{eq:trueerror}). The subscripts of $R$ and the arguments of $Acc$ are omitted when they are clear from the context.
	
	\begin{equation} \label{eq:truerisk}
		R_{\mathcal{D}, f^*}(h) = \underset{\mathnormal{x} \sim \mathcal{D}}{\mathbb{P}} [\mathnormal{h(x)\neq f^*(x)}] = \underset{\mathnormal{x} \sim \mathcal{D}}{\mathbb{E}} [\mathds{1}_{\mathnormal{h(x)\neq f^*(x)}}]
	\end{equation}
	
	\begin{equation}\label{eq:trueerror}
		{Acc}(\mathcal{D}, h) = 1 - R_{\mathcal{D}, f^*}(h) 
	\end{equation}

	Instead, the empirical risk, based on the training set $S$ (Equation \ref{eq:emprisk}) and the corresponding accuracy (Equation \ref{eq:empacc}) can be calculated:
	\begin{equation}
		\label{eq:emprisk}
		\widehat{R}_S(h)=\frac{1}{m} \: \sum_{i=1}^{m}{\mathds{1}_{h(x_i)\neq y_i}}
	\end{equation} 
	\begin{equation}
		\label{eq:empacc}
		\widehat{Acc}(S, h) = 1 - \widehat{R}_{S}(h)
	\end{equation}
	
	In theory, the estimation of the true risk is accurate since $\mathbb{E}[\widehat{R}_S(h)] = R(h)$. But because usually there are not arbitrary many samples in the training set $S$, it is not possible to confidently estimate $R(h)$ based on $S$. The learner is only able to choose a hypothesis based on $\widehat{R}_S(h)$. The learner might thus overfit, i.e. choose a hypothesis $h$, with $\widehat{R}_S(h) < R(h)$.

	To prevent overfitting, it is common to use three different datasets: (1) training set $S$, (2) validation set $V$, which is used to optimize hyperparameters, and (3) test set $T$, which is finally used to estimate the generalization. Correspondingly, $\widehat{Acc}(S)$, $\widehat{Acc}(V)$ and $\widehat{Acc}(T)$ can be measured and compared.

	Consequently, two types of overfitting are possible, as mentioned by Recht et al.~\cite{recht2018cifar}: training set overfitting and test set overfitting. The first is related to memorizing training dataset $S$. When a learner reaches $\widehat{R}_S(h) \approx 0$, this form of overfitting can be detected when a large difference between the training and the validation set accuracy is observed. Consequently, $\widehat{Acc}(S)-\widehat{Acc}(V)$ is commonly used to measure the overfitting of this type in practice. The reason for the test set overfitting is the gap between test set accuracy and the underlying data distribution. Currently, there is no uniform method to detect this type of overfitting, although the first attempts have been made~\cite{recht2018cifar,werpachowski2019detecting,recht2019imagenet}.

	\subsection{Related Work}
	\subsubsection{Perturbed Model Validation (PMV)} 
	
	Zhang et al.~\cite{zhang2019perturbed} mainly focused on selecting a model with the best fit from a set of candidate models based on a given dataset. The final model should neither overfit (and perform badly on unseen data) nor underfit (perform badly on all data), which would indicate a poorly chosen model. The authors recognized the weakness of traditional overfitting measures, such as cross-validation, VC-dimension, and Rademacher complexity, and proposed a new framework to evaluate the fit between the model and the data at hand. In PMV, the noise of different levels is injected into the data, followed by re-training the model on the perturbed data while measuring the decrease of the empirical accuracy (cf. Equation [\ref{eq:empacc}]). To quantify the results, linear regression is applied to the empirical accuracies for different noise levels $r_i$, yielding: 
	\begin{equation}
		\widehat{Acc}(S_{r_i}) = \widehat{Acc}(S_{0}) + k\cdot r_i
	\end{equation}
	The absolute value of the slope of the linear fit is then used as an overfitting measure:
	\begin{equation}
		\label{eq:PTV}
		\hat{k}_S = \left| \frac{\sum_{i=0}^{p} (r_i-\overline{r})( \widehat{Acc}(S_{r_i})- \overline{\widehat{Acc}(S_r))}}{\sum_{i=0}^{p}(r_i - \overline{r})^2} \right|,
	\end{equation} where $\overline{\widehat{Acc}(S_r)}$ is the mean accuracy over all perturbation degrees $r_i$. Thus, a higher value of $\hat{k}_S$ indicates a greater decrease rate and hence a better model fit for a given $S$.

	This way, PMV does not require splitting the available data and makes it possible to measure the level of overfitting. Furthermore, it facilitates selecting a model that fits the data distribution best. However, the authors discuss the application of PMV for neural networks very briefly, suggesting that PMV could be used for hyperparameter tuning. This is shown for a neural network on the MNIST dataset~\cite{deng2012mnist}. It remains unclear, whether the approach helps to perform model selection and detection of overfitting in CNNs.

	\subsubsection{Detection of Overfitting via Adversarial Examples} Werpachowski et al.~\cite{werpachowski2019detecting} claim, that the empirical risk on a test set as an estimator for the expected risk is significantly weakened if the learned hypothesis $f$ and the test set $T$ are not independent.
	To tackle this challenge, the authors constructed a new risk estimator that aims to be insensitive to test set overfitting. The new estimator is based on a perturbed test dataset and uses an adversarial generator. 
	
	In practice, the proposed procedure consists in generating a new set $T'=\{(x'_i,y'_i), i=1,..,m\}$ of $m$ adversarial examples using an adversarial example generator $g$. Then the importance weighted adversarial error rate $\widehat{R}_g(f)$ is computed as follows:
	\begin{equation}
		\label{eq:Rg}
		\widehat{R}_g(f) = \frac{1}{m} \: \sum_{i=1}^{m}{\mathds{1}_{f(x'_i)\neq y_i} h_g(x'_i)}
	\end{equation}
	
	Next, this new error estimator is compared to the empirical error rate $\widehat{R}_T(f)$. The hypothesis on the independence of $T$ and $h$ is then rejected, if the difference $\widehat{R}_g(f) - \widehat{R}_T(f)$  exceeds the empirical Bernstein bound.

	When applied to ImageNet classification task~\cite{russakovsky2015imagenet}, the proposed method has found no evidence of overfitting of the evaluated classifiers to the test set. The statement on the strong overfitting of the VGG-like architectures~\cite{simonyan2014vgg} against the CIFAR-10 test set~\cite{krizhevsky2009cifar}, which was present in the first version of the paper, was omitted in the final version of the paper.     
	
	\section{Concept}
	
	\subsection{Pool of Models}
	To ensure evaluation of the proposed approaches, we first define a pool of models with the predefined level of overfitting to the training data. For this, we use VGG~\cite{simonyan2014vgg} and ResNet~\cite{he2016resnet} models. Each of the architectures was altered in the setup and training routine, using features that are known to influence the overfitting behavior of neural networks. In total that yielded a pool of 16 models. 
	
	We focus on the image classification task; all models are trained on the CIFAR-10 dataset~\cite{krizhevsky2009cifar}. To stay consistent and be able to compare models, the same validation and test data were used for all models and approaches.

	Table \ref{tab:model_definitions} gives an overview of the techniques that we have introduced to create a spectrum of models with various overfitting behavior. These techniques fall into three categories: regularization, number of parameters, and size of the training set. In the following, we provide a summary of the applied (anti-)overfitting measures for VGG and ResNet. To assess the overfitting behavior, we measure the gap between training and validation accuracy $\widehat{Acc}(S)-\widehat{Acc}(V)$.

	\subsubsection{VGG}
	
	Since VGG was mainly designed for the ImageNet dataset, some adjustments were made to facilitate a successful application on CIFAR-10. Table \ref{tab:vgg_architecture} shows details of the architecture.

	The following regularization techniques were introduced to the VGG model to get models C1, C3, C5 and C7:
	\begin{itemize}
		\item batch normalization after each convolutional layer, as suggested by Liu et al.~\cite{liu2015vggcifar}. This introduces an inductive bias which helps to prevent overfitting. 
		\item $l_2$ weight regularization with a factor of $\lambda = 0.005$, which prevents the layer weights to get too large. 
		\item multiple dropout units with an average dropout of $0.3$.
		\item normalizing the input data to $[0,1]$.
		\item larger batch size of 128.
		\item training with the SGD algorithm with a momentum of $0.9$ and an adaptive learning rate, starting from $0.1$ with a decay of $0.000001$. A high learning rate in the beginning prevents the risk to get stuck in a local minimum. After multiple epochs, the learning rate is decreased to allow convergence towards a found optimal minimum. 
		\item data augmentation, including small transformations like shifting, rotation, and horizontal flips. This improves generalization since the model also learns from data that differ from the images in the training set.  
	\end{itemize}
	
	The unregularized models C2, C4, C6, C8 do not use batch normalization, weight regularization, and dropout, the pixel values are kept in $[0,255]$, the batch size is 32.  The smaller batch size results in a weak estimation of the general gradients, which weakens the generalization performance of the model. The training was performed with the SGD algorithm without momentum and a very small, fixed learning rate of $0.0001$. The small learning rate intentionally increases the risk to converge towards a local optimum in the beginning.
	
	To get models with more parameters (C3, C4, C7, and C8), we have introduced two fully connected layers before the output neurons (see Table \ref{tab:vgg_architecture}). The remaining models C1, C2, C5, and C6 have only a single fully connected layer -- this is sufficient to combine the features that are extracted from the CIFAR-10 data. 
	
	For models, trained on a larger dataset C1-C4, the available data were split with 90\% to be used for training (i.e. 45,000 images), while 10\% (i.e. 5,000 images) are used for validation. While it is still much less than the training images from ImageNet, the amount seems to be sufficient to learn the necessary patterns of the CIFAR-10 dataset. The remaining models C5-C8 were trained on 5,000 images, sampled from the larger dataset.
	
	\begin{table}[t]
		\centering
		\caption{Model definitions}
		\label{tab:model_definitions}
		\begin{tabular}{|c |l|c|c|c|c|}
			\hline
			No & Name   & Reg. 	& \# Params  & Training set \\ \hline
			C1 & Regularized   & yes   	 & small	& 45 K     \\ 
			C2 & Unregularized   & no & small  & 45 K      \\
			C3 & Regularized fat   & yes & large & 45 K    \\  
			C4 & Unregularized fat   & no & large & 45 K      \\ \hline
			C5 & Regularized, less data  & yes   	 & small	& 5 K     \\ 
			C6 & Unregularized, less data   & no & small & 5 K      \\
			C7 & Regularized fat, less data   & yes & large & 5 K      \\
			C8 & Unregularized fat, less data   & no & large & 5 K      \\\hline
		\end{tabular}
	\end{table}
	
	Table \ref{tab:vgg_performance} shows performance of baseline VGG models. According to $\widehat{Acc}(S)-\widehat{Acc}(V)$, we observe the rising overfitting level from model C1 to C8. In particular, for models C1-C3, the performance on the training data is close to the performance on the validation and test data, meaning that the model does not overestimate its performance based on the training data and is therefore not overfitted against the training data. On the other hand, for models C4-C8 the performance on the training data differs much from the performance on the validation and test data, meaning that the model does over-estimate its performance based on the training data and is therefore overfitted against the training data. 
	
	\begin{table}[t]
		\caption{VGG Architecture: n $(m\times m)$-conv is a convolutional layer with n $m\times m$ filters, $n$ FC-layer is a fully-connected layer with n neurons. $\times n$ means the layer is repeated $n$ times.}
		\label{tab:vgg_architecture}
		\centering
		\begin{tabular}{|c|c|}
			\hline
			\textbf{VGG }         & \textbf{VGG Fat} \\ \hline
			64 $(3\times 3)$-conv  & 64 $(3\times 3)$-conv   \\
			64 ($(3\times 3)$-conv  & 64 $(3\times 3)$-conv  \\
			(2x2) max-pooling            & (2x2) max-pooling  \\
			128 $(3\times 3)$-conv & 128 $(3\times 3)$-conv \\
			128 $(3\times 3)$-conv   & 128 $(3\times 3)$-conv \\
			(2x2) max-pooling     & (2x2) max-pooling \\
			256 $(3\times 3)$-conv $\times$ 3 & 256 $(3\times 3)$-conv $\times$ 3 \\
			(2x2) max-pooling            & (2x2) max-pooling  \\
			512 $(3\times 3)$-conv $\times$ 3 & 256 $(3\times 3)$-conv $\times$ 3\\
			(2x2) max-pooling            & (2x2) max-pooling \\
			512 $(3\times 3)$-conv $\times$ 3 & 256 $(3\times 3)$-conv $\times$ 3 \\
			512 FC-layer & 4096 FC-layer  \\
			& 4096 FC-layer \\ \hline
		\end{tabular}
	\end{table}
	
	\begin{table}[t]
		\centering
		\caption{Accuracy of baseline VGG models}
		\label{tab:vgg_performance}
		\begin{tabular}{| c |c|c|c|c|c|}
			\hline
			Model & $\widehat{Acc}(S)$   	& $\widehat{Acc}(V)$  & $\widehat{Acc}(T)$  & $\widehat{Acc}(S)-\widehat{Acc}(V)$ \\ \hline
			C1 &  98.65 & 91.7	& 91.5 & 6.95 \\ 
			C2 &  99.98 & 70.62  & 70.70 & 29.36    \\
			C3 &  99.99 & 89.92 & 89.52 & 10.07   \\  
			C4 &  100.00 & 77.22 & 76.05 & 22.78    \\ 
			C5 &  99.77 & 69.06 & 68.99 & 30.71   \\ 
			C6 &  99.94 & 54.32 & 55.00 & 45.62     \\
			C7 &  100.00 & 57.54 & 58.75 & 42.26    \\
			C8 &  100.00 & 45.80 & 46.00 & 54.2 0   \\\hline
		\end{tabular}
	\end{table}
	
	\subsubsection{ResNet}
	Our architecture of ResNet~\cite{he2016resnet} is inspired by the Keras implementation~\cite{chollet2015keras} of ResNet with a depth of 20 layers which are made up of 6 residual blocks. Table \ref{tab:resnet_architecture} shows details of the architecture. In the following, the difference between models C1-C8 is explained.
	
	\begin{table}[t]
		\caption{ResNet Architecture:  ResBlock $n$ is a residual block with $n$ filters per layer, $n$ FC-layer is a fully-connected layer with n neurons. $\times n$ means the block is repeated $n$ times.}
		\label{tab:resnet_architecture}
		\centering
		\begin{tabular}{|c|c|}
			\hline
			\textbf{ResNet} & \textbf{ResNet Fat} \\ \hline
			ResBlock 16 $\times$ 3 & ResBlock 16 $\times$ 7 \\
			ResBlock 32 $\times$ 3 & ResBlock 32 $\times$ 7 \\
			ResBlock 64 $\times$ 3 & ResBlock 64 $\times$ 7 \\
			(8x8 pool) & (8x8 pool) \\
			10 FC-layer & 10 FC-layer \\ \hline
			
		\end{tabular}
	\end{table}
	
	\begin{table}[t]
		\centering
		\caption{Accuracy of baseline ResNet models}
		\label{tab:resnet_performance}
		\begin{tabular}{|c|c|c|c|c|c|c|}
			\hline
			Model  &$\widehat{Acc}(S)$	& $\widehat{Acc}(V)$ & $\widehat{Acc}(T)$ & $\widehat{Acc}(S)-\widehat{Acc}(V)$ \\ \hline
			C1 &  98.67 & 90.98 & 90.31 & 7.69 \\ 
			C2 &  91.05 & 82.78  & 08.54 & 8.27      \\
			C3 & 100 & 87.92 & 87.34 & 12.08   \\  
			C4 & 99.05 & 82.26 & 82.12 & 16.79  \\
			C5 & 99.98 & 51.26 & 51.95 & 48.72   \\ 
			C6 & 97.73 & 53.64 & 53.36 & 44.09   \\
			C7 & 99.98 & 50.40 & 51.06 & 49.58   \\
			C8 & 99.88 & 54.34 & 54 & 45.54   \\\hline
		\end{tabular}
	\end{table}
	
	The following regularization techniques were introduced to the ResNet architecture to get models C1, C3, C5, and C7:
	\begin{itemize}
		\item batch normalization applied to every layer in the network. 
		\item $l_2$ weight regularization with a factor of $\lambda = 0.005$.
		\item training with the Adam optimizer~\cite{kingma2014adam} with an adaptive learning rate. 
		\item normalizing the input to $[0,1]$.
		\item larger batch size of 128.
		\item data augmentation with shifting, translating, and horizontal flipping.
	\end{itemize}

	Dropout was not used to increase the performance of regularized models. This is motivated by the following facts: (1) the ResNet architecture renounces fully connected layers (besides the last output layer) and hence has a smaller fraction of weights that are learned, compared to VGG, and (2) ResNet heavily relies on batch normalization to regularize the weights and introduce some inductive bias. The combination of dropout and batch normalization weakens the performance of many architectures, as shown by Li et al.~\cite{li2019dropoutBN}.
	
	The unregularized ResNet models were trained with the Adam optimizer but with a fixed learning rate of $0.0001$. During the experiments, this amplified the desired overfitting behavior for the ResNet architecture. The training lasted for 250 epochs as in the regularized case, but a smaller batch size of 16 was used and no weight regularization was applied. The input data were not normalized and stayed in the range of $[0,255]$. Techniques like data augmentation and batch normalization were not applied either. 
	
	Table \ref{tab:resnet_performance} shows performance of baseline ResNet models. Similar to VGG, the gradual increase in $\widehat{Acc}(S)-\widehat{Acc}(V)$ can be observed from C1 to C8.
	
	\subsection{Perturbation Methods}
	
	Table \ref{tab:methods_data} gives an overview of the proposed approaches to detect overfitting against the training data. While some methods manipulate the training process of a neural network, others use a pre-trained model and evaluate its accuracy on modified training datasets. 
	
	\begin{table}[t]
		\centering
		\caption{Overview of the perturbation methods for measuring overfitting to the training dataset.}
		\label{tab:methods_data}
		\begin{tabular}{|l|l|l|}
			\hline
			\textbf{Method} 	& \textbf{Training Data} & \textbf{Test Data}  \\ \hline
			PMV	& Perturbed (label noise) & Perturbed (label noise) \\ \hline
			PTV & Perturbed (label noise) & Original\\ \hline
			FGSM				& Original &Perturbed (FGSM)     \\ \hline
			Spatial Attack      	& Original & Perturbed (shift, rotation)  \\ \hline
			Gaussian Noise     		& Original & Perturbed (Gaussian) \\ \hline
			Corruptions	& Original & Perturbed  (corruptions)\\ \hline
		\end{tabular}
	\end{table}

	\subsubsection{Perturbed Model Validation (PMV)} Following the procedure described by Zhang et al.~\cite{zhang2019perturbed}, we retrain each model several times with an increased degree of label noise. The authors of PMV suggest randomly exchanging labels for a portion of data. To ensure that the newly assigned labels are false, we assign a random, false label to a portion $r$ of the training data, whereas $r \in \{0,0.1,0.2,0.3,0.4,0.5\}$. For each training a new, untrained (randomly initialized) version of the model is used, so that previous training runs do not influence the current training. After each training, the accuracy is evaluated on the perturbed data that were used for training.
	
	\subsubsection{Perturbed Training Validation (PTV)} follows the same training procedure as in PMV, but the accuracy is measured on the unperturbed training set. A model with a better fit is expected to yield better accuracy on the original training data because it has learned the true patterns, despite the noise during training. On the other hand, we expect an overfitted model to perform worse since it fits the perturbed labels. Our intuition is supported by the work by Rolnick et al.~\cite{rolnick2017deeprobust}, which states that neural networks manage to learn the underlying true relations, even in data with a very high level of noise.
	
	\begin{figure*}[t]
		\centering
		\includegraphics[width=0.49\linewidth]{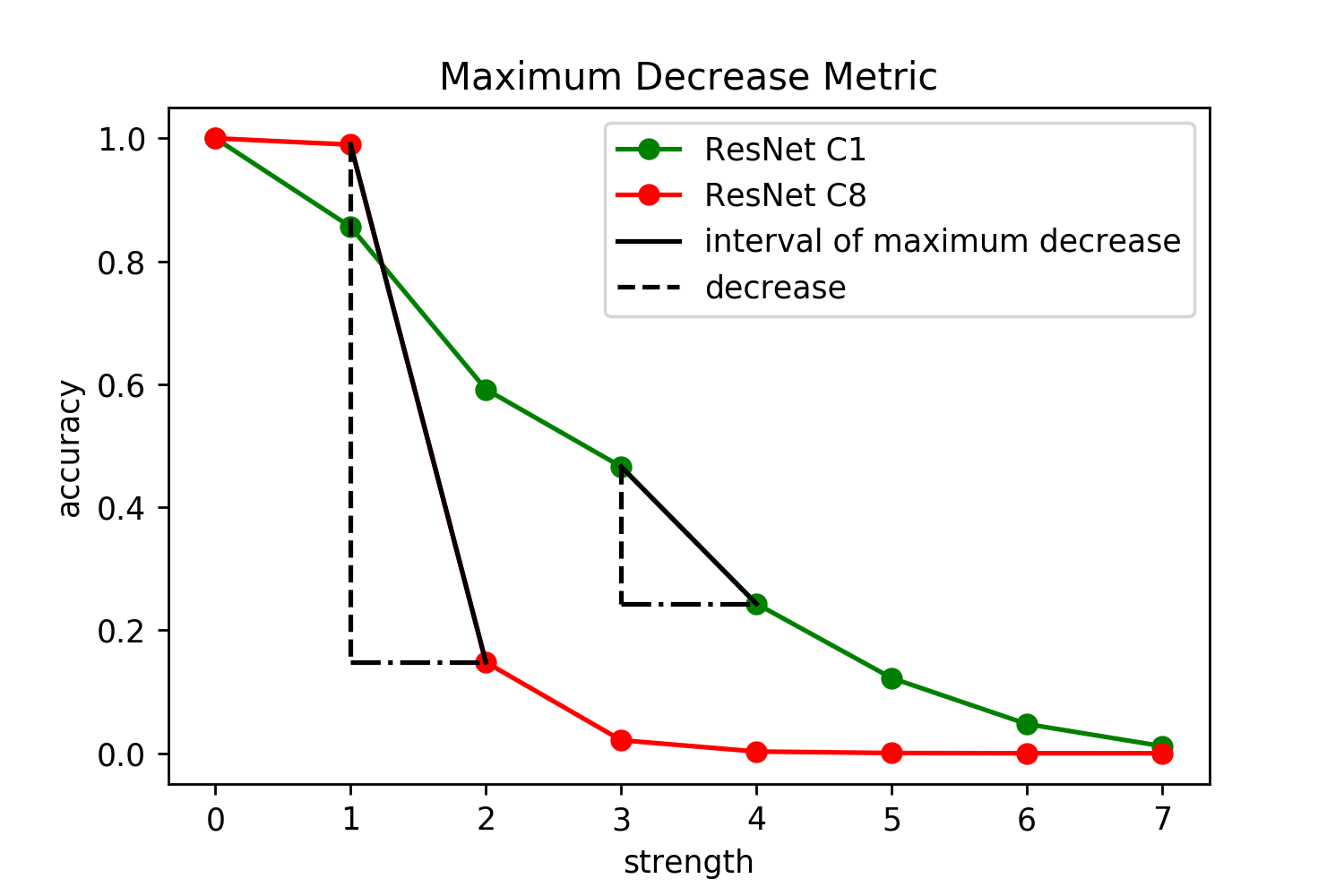}
		\includegraphics[width=0.49\linewidth]{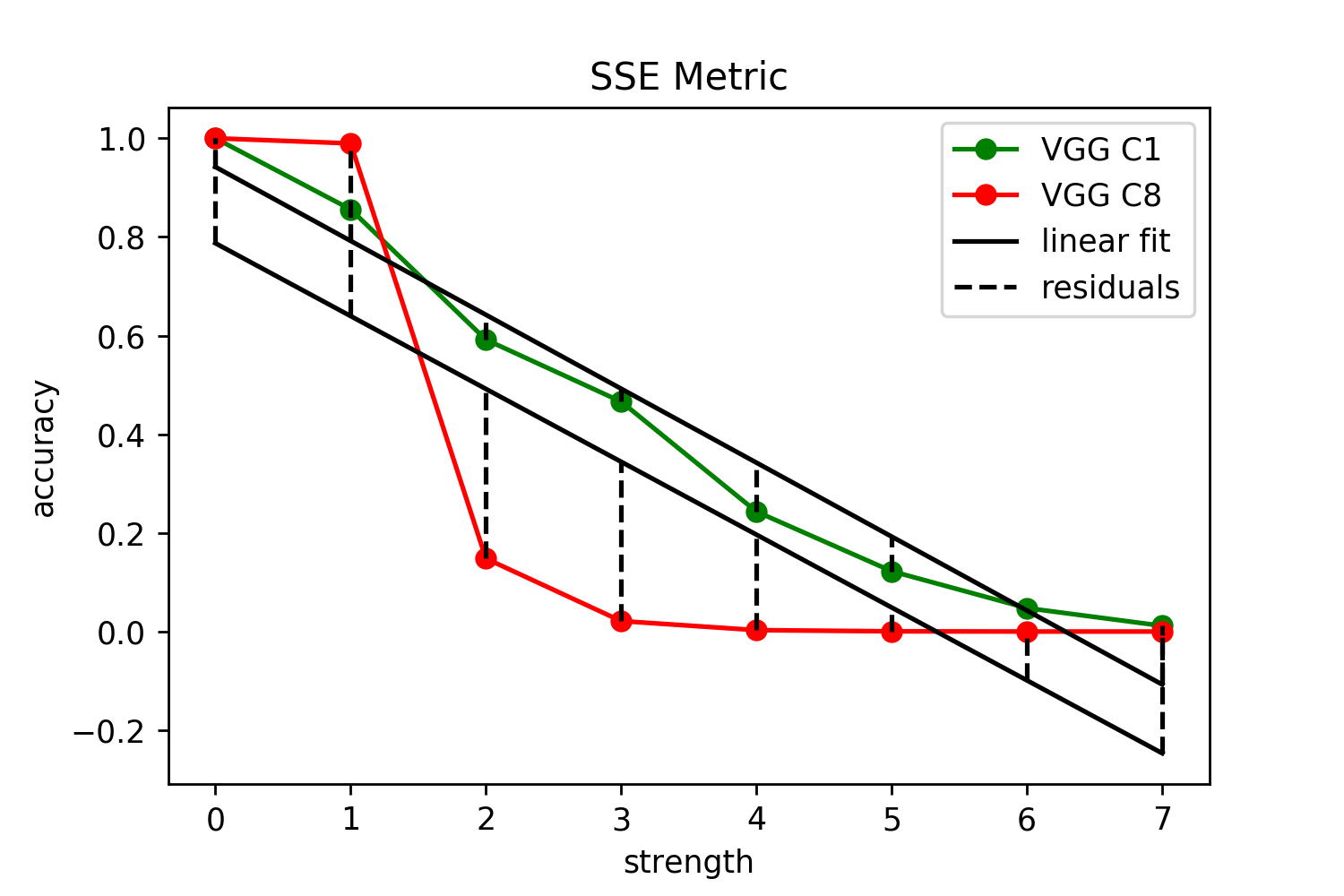}
		\caption{Left: maximum decrease metric, exemplary for the ResNet C1 and C8 models. Right: SSE metric, exemplary for the VGG C1 and C8 models.}	
		\label{fig:sse}
	\end{figure*}
	
	PTV is an improvement over PMV for several reasons: (1) unlike PMV, it returns an approximate estimation of $R(h)$, (2) PTV can be used with very few (or even only one) value for $r$, which saves computational resources, and (3) PTV  empirically presents the weakness of overfitted models, which perform badly on data that have not been used during training.

	\subsubsection{FGSM Attack} Differently from the work by Werpachowski et al.~\cite{werpachowski2019detecting}, we focus on overfitting to the training set rather than to the test set. For this, we use white-box Fast Gradient Sign Method (FGSM)~\cite{goodfellow2014explaining} to generate adversarial perturbations. Although stronger variants of FGSM like PGD~\cite{madry2017towards} exist, FGSM is a fast attack usually recommended for benchmarking. In FGSM, an adversarial example $x_{adv}$ is calculated by adding the weighted sign of the gradient of the loss function with respect to the input data:
	\begin{equation}
		\mathnormal{x_{adv} = x + \epsilon * sign(\nabla_x \mathcal{L}(\theta, {x}, y))}
	\end{equation}

	In~\cite{werpachowski2019detecting}, FGSM has led to impressive results on artificial examples. However, the authors could not quantify the distribution shift for image data, so FGSM could not be applied to image classifiers. Since our goal is not to estimate $R(f)$ as in~\cite{werpachowski2019detecting}, we use FGSM on image classifiers and observe differences between overfitted and correct models. Differently from the methods mentioned above, we do not retrain the model on perturbed data but evaluate the performance. 
	
	\subsubsection{Spatial Attacks} CNNs have been shown to be vulnerable to small image translations or rescalings~\cite{azulay2018smalltranslation}. To be able to calculate $h_g(x)$ in~\cite{werpachowski2019detecting}, it was necessary to use random shifting instead of targeted shifting to fool the model. This constraint does not apply if the main goal is to measure overfitting. We use the spatial attack for image data as defined by Engstrom et al.~\cite{engstrom2019exploring}. To apply different strengths of the attack, the attack strength metric $\alpha$ is used. This allows the shifting of an image by up to $\alpha$ pixels in each direction and the rotation of an image by up to $\alpha$ degrees. 
	
	The attacks for each model were specifically chosen, based on multiple rotations and shifting attempts. In contrast to the FGSM attack, this approach measures overfitting without access to the model gradients, since it is based solely on altering the input and observing the output.

	\subsubsection{Gaussian Noise}
	While the FGSM attack and spatial attacks are targeted specifically at one model, it is interesting to observe the behavior of the models on random noise. 
	Usually, random noise is used to measure the corruption robustness of a model instead of the robustness against targeted attacks. For example, Gilmer et al.~\cite{ford2019adversarialnoise} state, that the effects of adversarial examples and corruption noise are closely related. For our evaluation, we have used additive Gaussian noise. In particular, a Gaussian-distributed tensor in the shape of the image was generated and added to the original image, formally:
	\begin{equation}
		x_{gauss} = x + \epsilon * n,\text{  with } n \sim \mathcal{N}_{h\times w \times c}(\mathbf{0}_{h\times w \times c}, \mathbf{I}_{h\times w \times c})
	\end{equation} 
	with the zero tensor $\mathbf{0}$ for the mean, and the identity tensor $\mathbf{I}$ for the variance, which both have the image dimensions ${h\times w \times c}$. 
	Like the three previous methods, the perturbation was only applied to evaluate the performance of a pre-trained model, not for retraining the model itself. 
	
	\subsubsection{Hendrycks Corruptions} The Hendrycks framework for multiple corruptions~\cite{hendrycks2019benchmarking} is an established set of image modifications that are likely to occur due to changes in weather settings, camera conditions, or computational processing of the images. The corruptions fall into four main categories: noise, blur, weather, and digital. We apply the Hendrycks transformations to the CIFAR-10 training data and measure the accuracy of every model on the resulting corrupted datasets.

	\subsection{Overfitting Metrics}
	For PMV, we calculate the absolute value of the slope of the linear fit $\hat{k}_S$ as defined in Equation [\ref{eq:PTV}] and proposed by Zhang et al.~\cite{zhang2019perturbed}. Similarly to $\hat{k}_S$, we define the average decrease rate $\hat{k}^{PTV}_S$ for PTV. Differently from PMV, however, a large $\hat{k}^{PTV}_S$ value indicates overfitting since the true performance is strongly influenced by the noise degree during training. 
	
	We introduce two further metrics to measure the overfitting of a hypothesis $h$ to the training set $S$: the maximum decrease $\hat{k}_{max}(S,h)$ and the sum of squared errors $\widehat{SSE}(S,h)$ (see Figure \ref{fig:sse} for illustration).
	
	\begin{table*}[t]
		\centering
		\caption{Results of VGG models for different overfitting metrics}
		\label{tab:vgg_metrics}
		\begin{tabular}{|l |c|c|c|c|c|c|c|c| }
			\hline
			Approach & PMV & PTV & Spatial & Spatial & FGSM & FGSM & Gaussian & Gaussian \\ \hline
			Metric & $\hat{k}_S$ & $\hat{k}^{PTV}_S$ & $\hat{k}_{max}(S,h)$ & $\widehat{SSE}$ & $\hat{k}_{max}(S,h)$ & $\widehat{SSE}$ & $\hat{k}_{max}(S,h)$ & $\widehat{SSE}$ \\ \hline
			C1 & 0.9438 & \textbf{0.3481} &  \textbf{0.2642} & \textbf{0.0396} & 0.2795 & 0.1699 & 0.5315 & 0.7806\\ 
			C2 & 0.0015 & 0.9714  & 0.5932 & 0.3457 & 0.2758 & 0.6320 & 0.2980 & 0.6818  \\
			C3 & 1.0487 & 1.0934 & 0.2857 & 0.1070  & \textbf{0.0988} & \textbf{0.0215} & 0.3112 & 0.4456\\  
			C4 & \textbf{1.4231} & 0.8710 & 0.6146 & 0.3085 & 0.2487 & 0.0385 & 0.2432 & 0.6487\\ \hline
			C5 & 1.0849 & 0.9073 & 0.5102 & 0.2567 & 0.2762 & 0.0625 & 0.3234 & 0.7879\\ 
			C6 & 0.0405 & 0.9959 & 0.7268 & 0.4248 & 0.6030 & 0.3749 & 0.2986 & 0.6558  \\
			C7 & 0.9927 & 1.0345 & 0.4966 & 0.3597 & 0.3410 & 0.3314 & \textbf{0.0622} & \textbf{0.0204}\\
			C8 & 0.0023 & 0.9996 & 0.8416 & 0.5007 & 0.4546 & 0.0676 & 0.1208 & 0.0417\\\hline
		\end{tabular}
	\end{table*}
	
	\begin{table*}[t]
		\centering
		\caption{Results of ResNet models for different overfitting metrics}
		\label{tab:resnet_metrics}
		\begin{tabular}{|l |c|c|c|c|c|c|c|c|}
			\hline
			Approach & PMV & PTV & Spatial & Spatial & FGSM & FGSM & Gaussian & Gaussian\\ \hline
			Metric & $\hat{k}_S$ & $\hat{k}^{PTV}_S$ & $\hat{k}_{max}(S,h)$ & $\widehat{SSE}$ & $\hat{k}_{max}(S,h)$ & $\widehat{SSE}$ & $\hat{k}_{max}(S,h)$ & $\widehat{SSE}$ \\ \hline
			C1 & 0.9654 & 0.4374 & 0.3875 & 0.1687 & 0.1832 & 0.3029 & 0.3519 & 0.5859\\ 
			C2 & 0.6087 & 0.9229  & \textbf{0.2542} &\textbf{0.0810} & 0.1789 & 0.3493 & 0.2987 & \textbf{0.4821} \\
			C3 & \textbf{0.9942} & \textbf{0.3737} & 0.3180 & 0.0873 & 0.2598 & 0.4353 & 0.3886 & 0.6115\\  
			C4 & 0.1295 & 0.9925 & 0.3889 & 0.1749 & \textbf{0.1601} & \textbf{0.2925} & 0.3778 & 0.7452\\ \hline
			C5 & 0.4505 & 0.9186 & 0.4900 & 0.3876 & 0.3432 & 0.8625 & 0.3350 & 0.5964\\ 
			C6 & 0.0280 & 1.0078 & 0.7528 & 0.4233 & 0.2956 & 0.3149 & 0.2566 & 0.5374\\
			C7 & 0.0002 & 0.9998 & 0.5950 & 0.4030 & 0.3380 & 0.8391 & 0.3528 & 0.6499\\
			C8 & 0.0237 & 0.9984 & 0.5360 & 0.3770 & 0.2368 & 0.4468 & \textbf{0.2254} & 0.4988\\\hline
		\end{tabular}
	\end{table*}
	
	\subsubsection{Maximum Decrease Metric.} $\hat{k}_{max}(S,h)$ is calculated as the maximal decrease of accuracy within an interval of one:
	\begin{equation}
		\hat{k}_{max}(S,h) = \max \: |\widehat{Acc}(S_r) - \widehat{Acc}(S_{r-1})|, r = 1,..,p
	\end{equation}
	A high and sudden drop in accuracy seems to represent an overfitted model. Therefore, a higher value of the maximum decrease metric $\hat{k}_{max}(S,h)$ would mean a higher degree of overfitting against the dataset $S$.

	\subsubsection{Sum of Squared Errors (SSE) Metric.} Motivated by the fact, that the models with correct fit mostly exhibit linear behavior, we define $\widehat{SSE}(S,h)$ as the squared sum of the residuals between the observed accuracies and the best linear fit:
	\begin{equation}
		\widehat{SSE}(S,h) = \sum_{r=1}^{p} (\widehat{Acc}(S_r)- (\widehat{Acc}(S_0) + k * r))^2
	\end{equation}
	A higher value of $\widehat{SSE}(S,h)$ indicates a worse linear approximation and hence a higher degree of overfitting.

	\section{Experiments and Evaluation}
	
	\subsection{Results for the Proposed Methods}
	
\begin{figure*}[hbtp]
		\centering
		\includegraphics[width=0.37\linewidth]{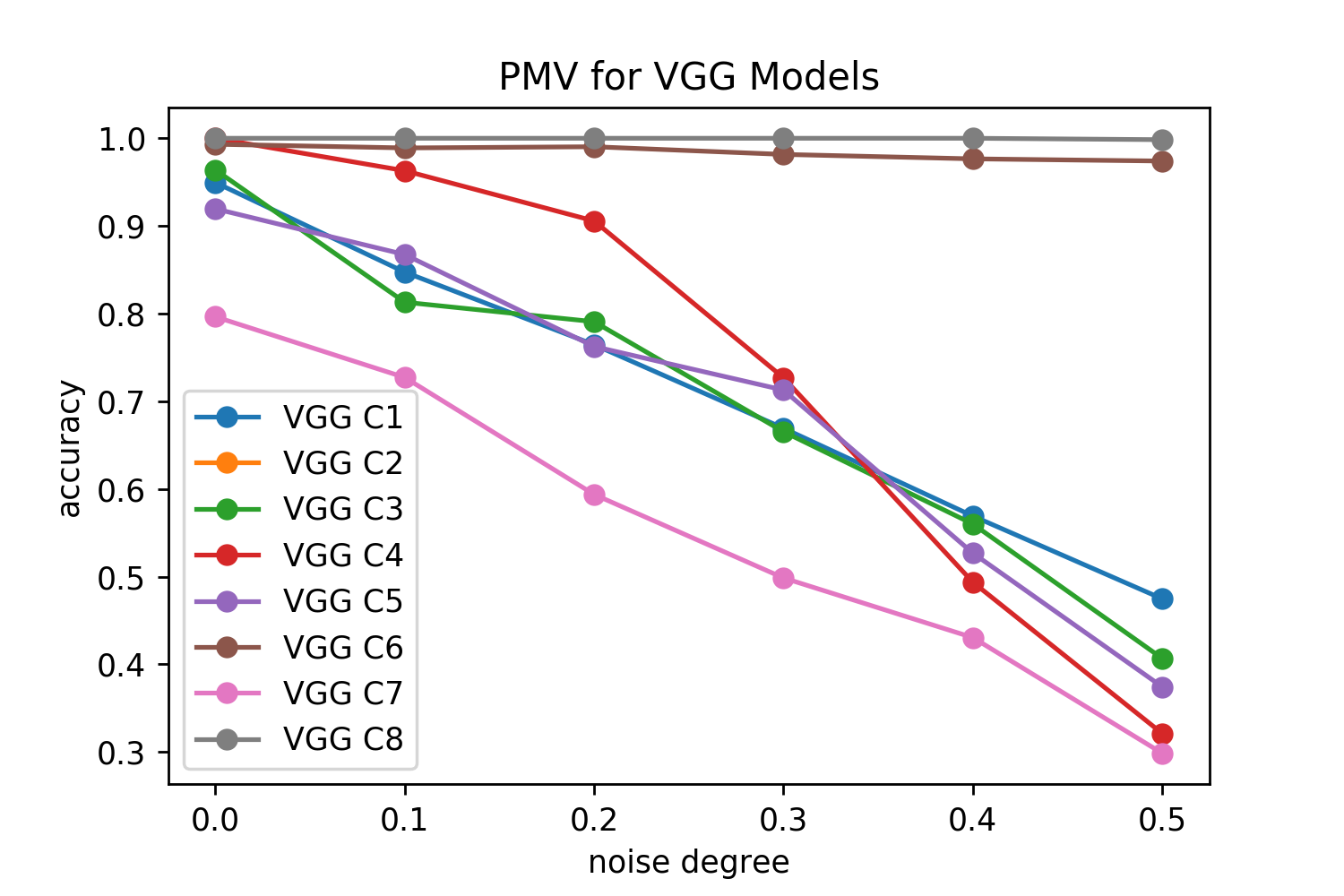}
		\includegraphics[width=0.37\linewidth]{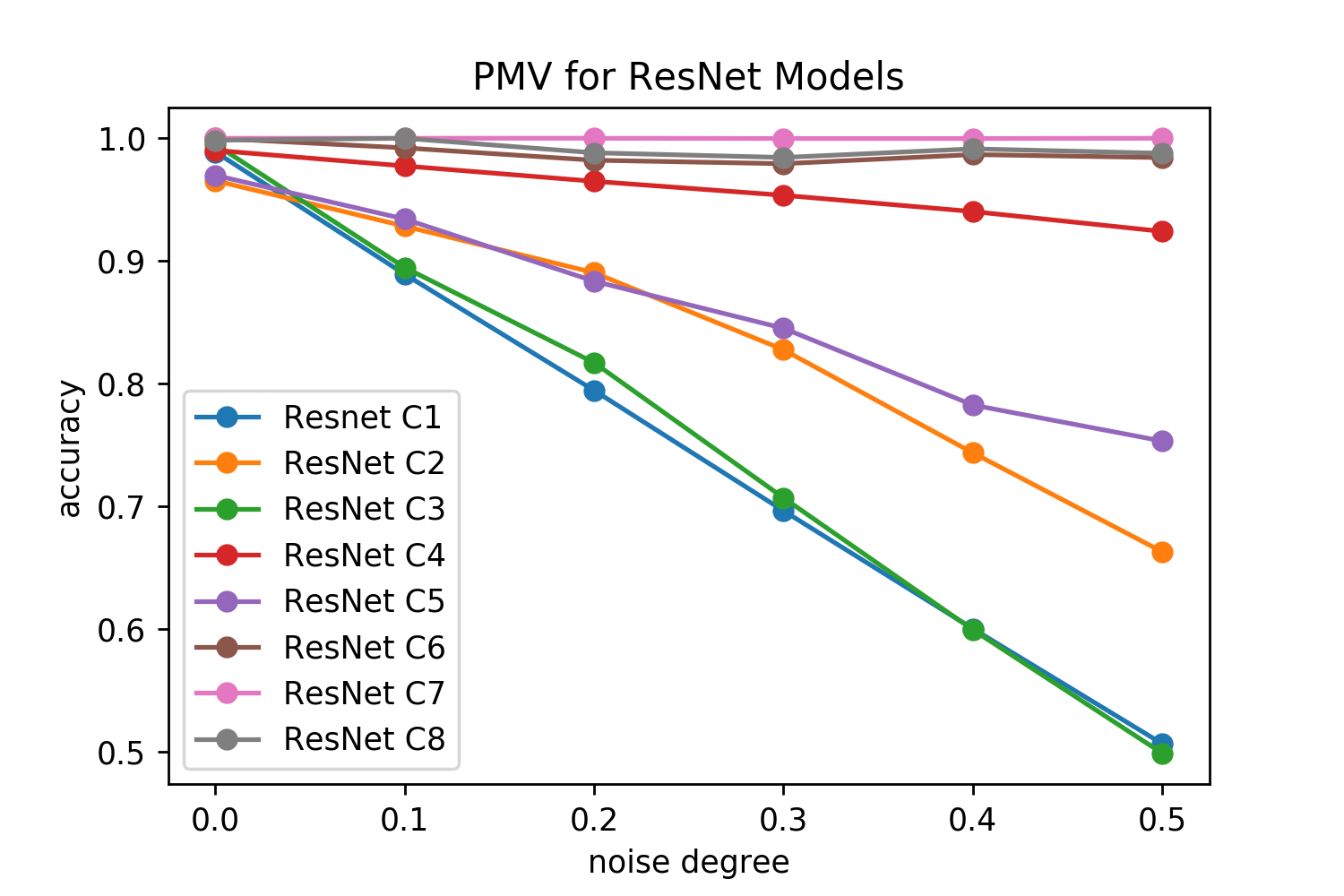}\\
		\includegraphics[width=0.37\linewidth]{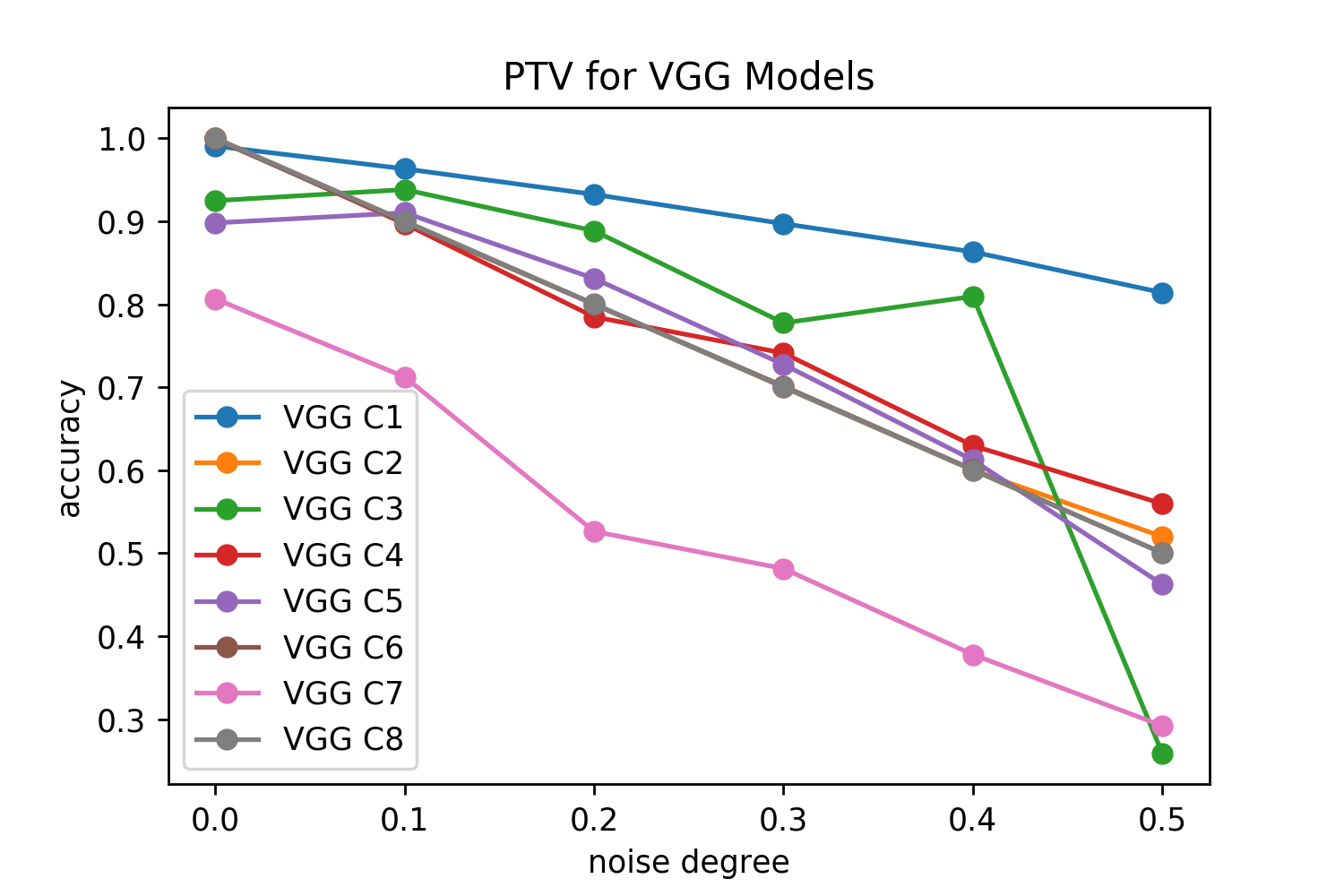}
		\includegraphics[width=0.37\linewidth]{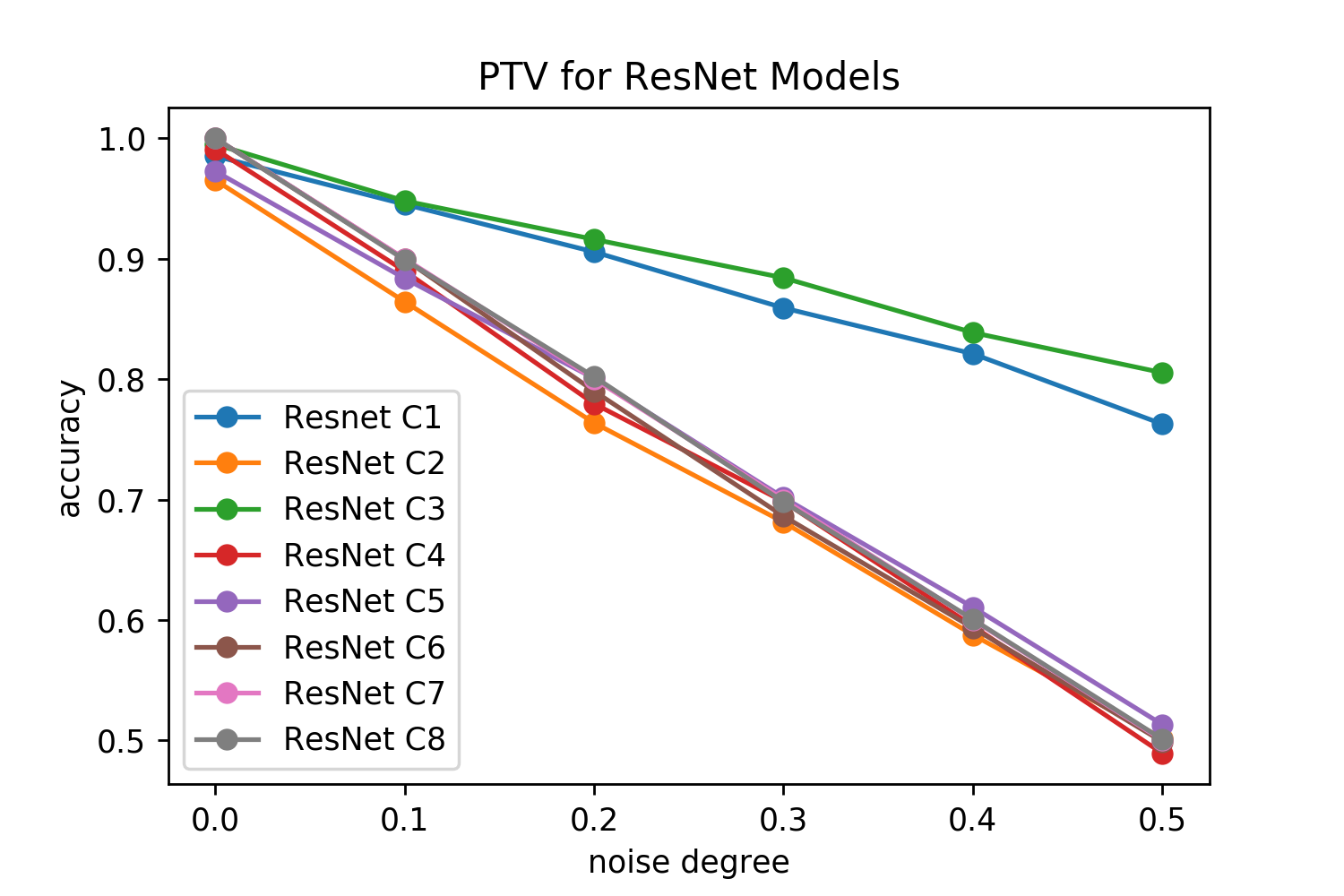}\\
		\includegraphics[width=0.37\linewidth]{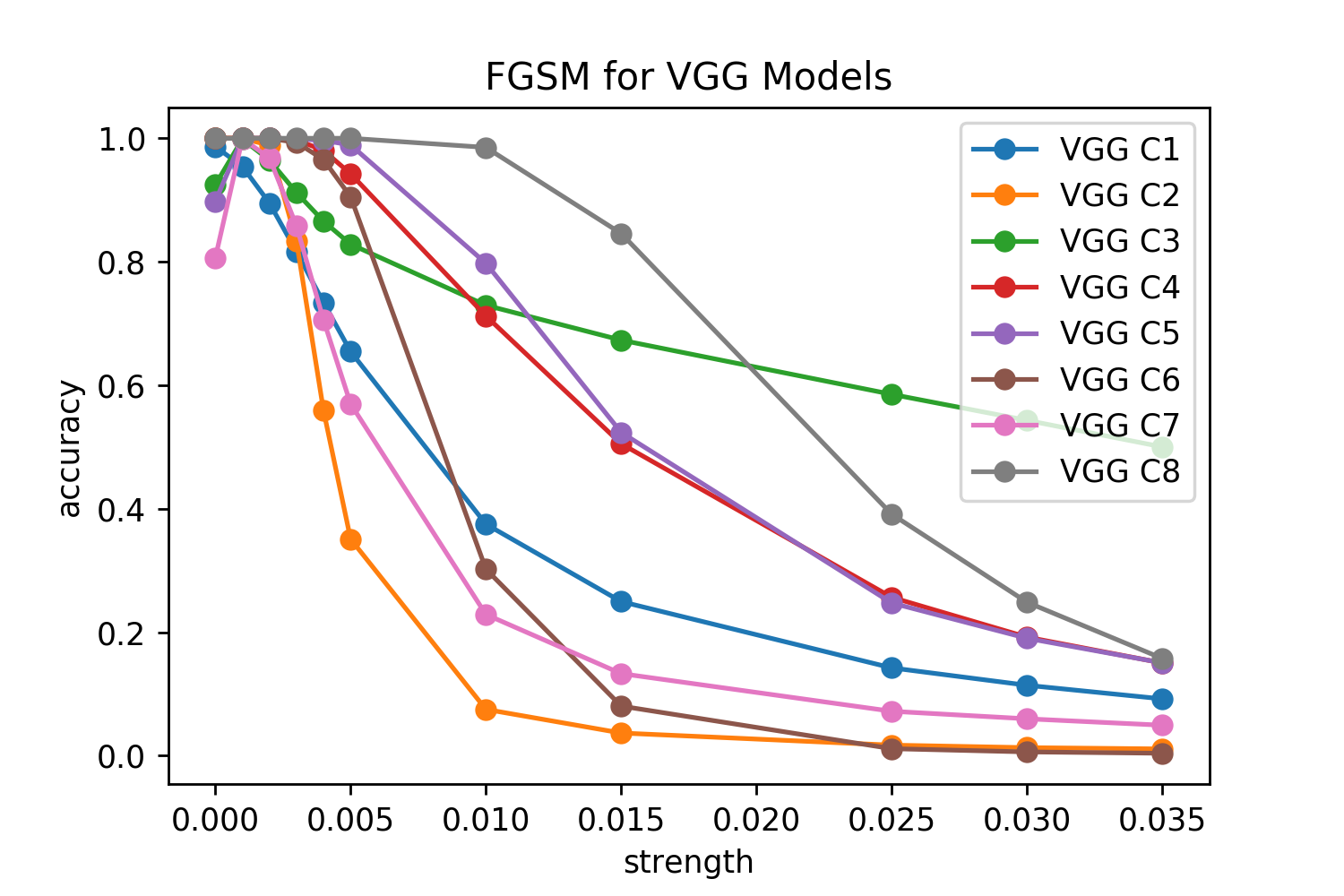}
		\includegraphics[width=0.37\linewidth]{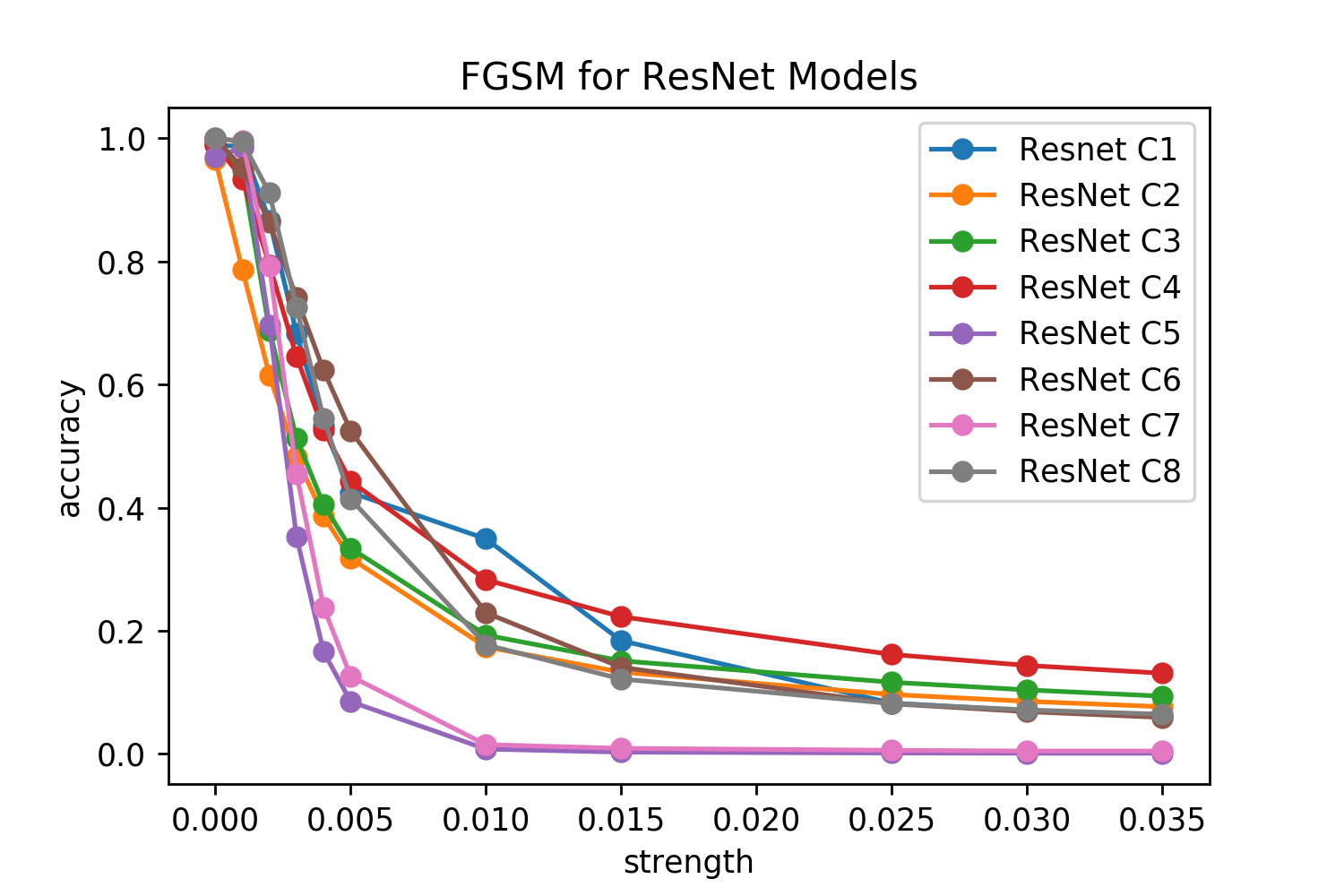}\\
		\includegraphics[width=0.37\linewidth]{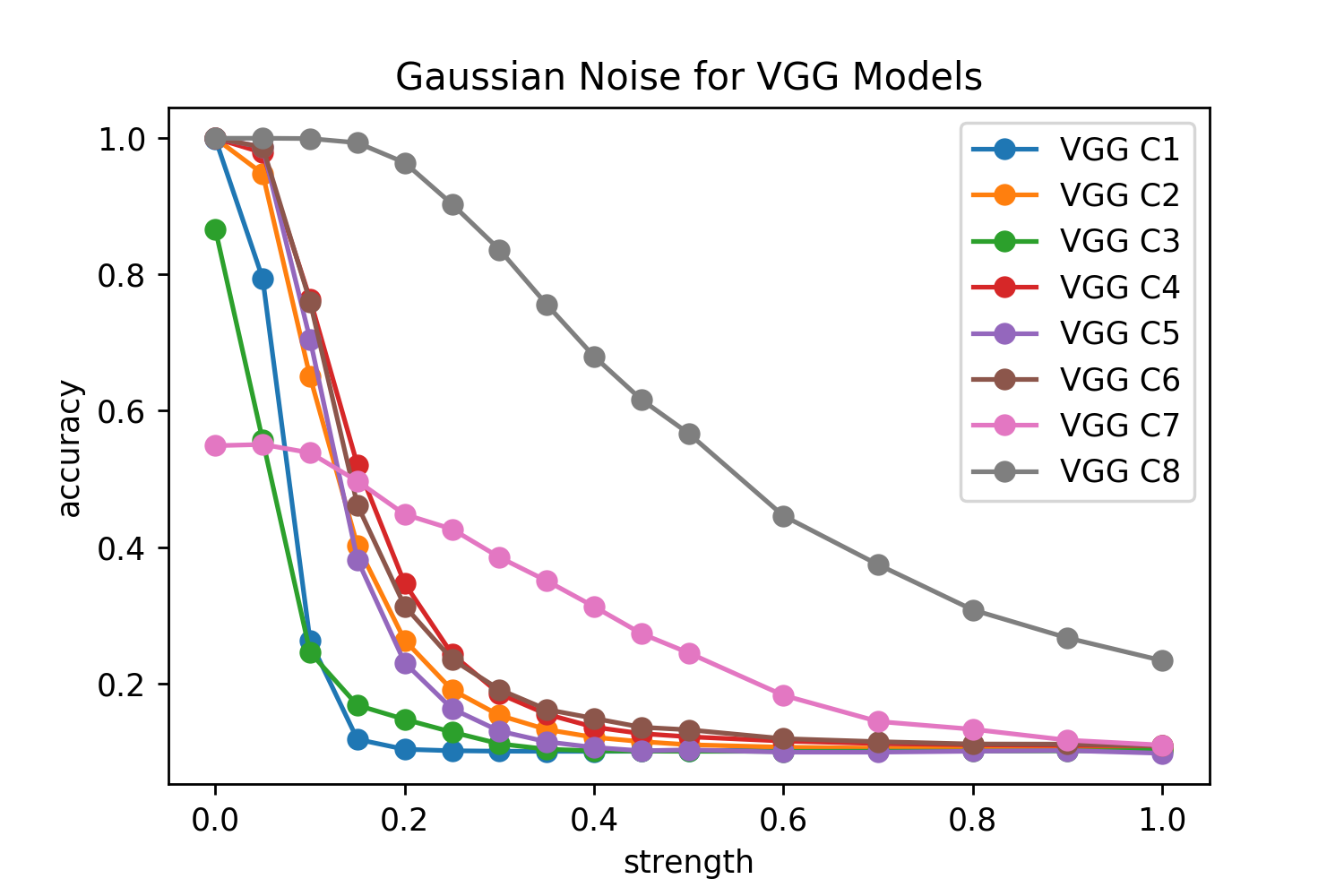}
		\includegraphics[width=0.37\linewidth]{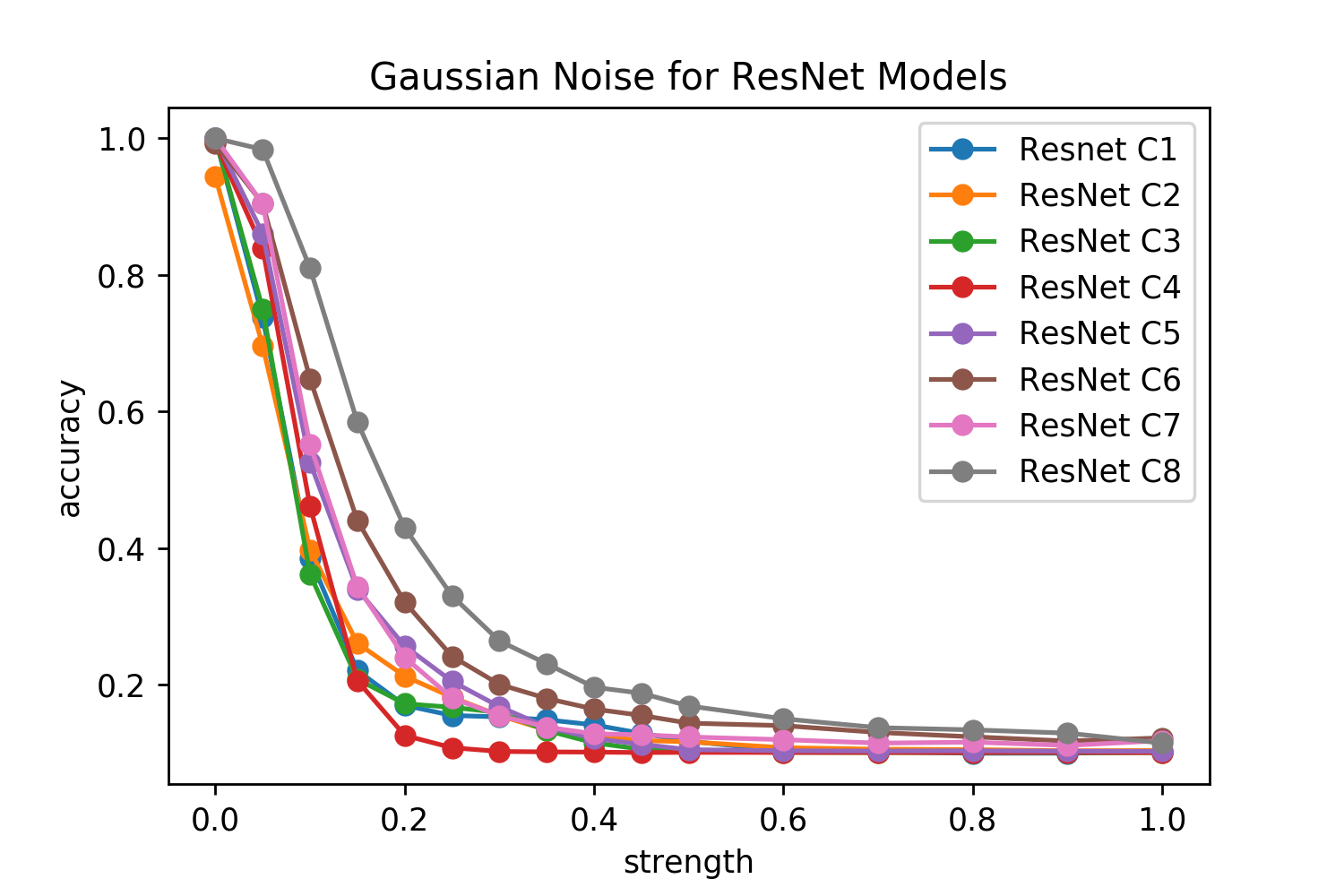}\\
		\includegraphics[width=0.37\linewidth]{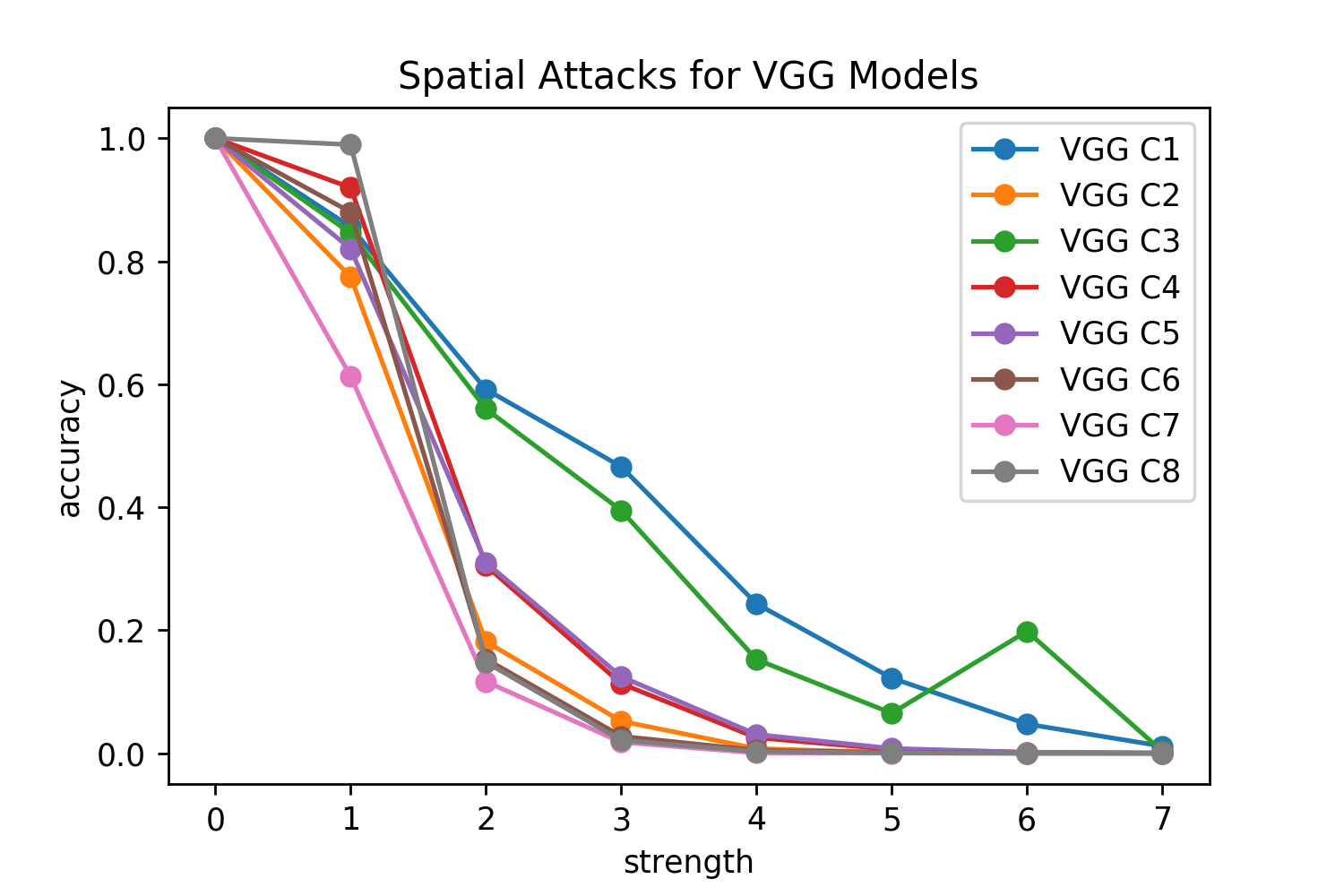}
		\includegraphics[width=0.37\linewidth]{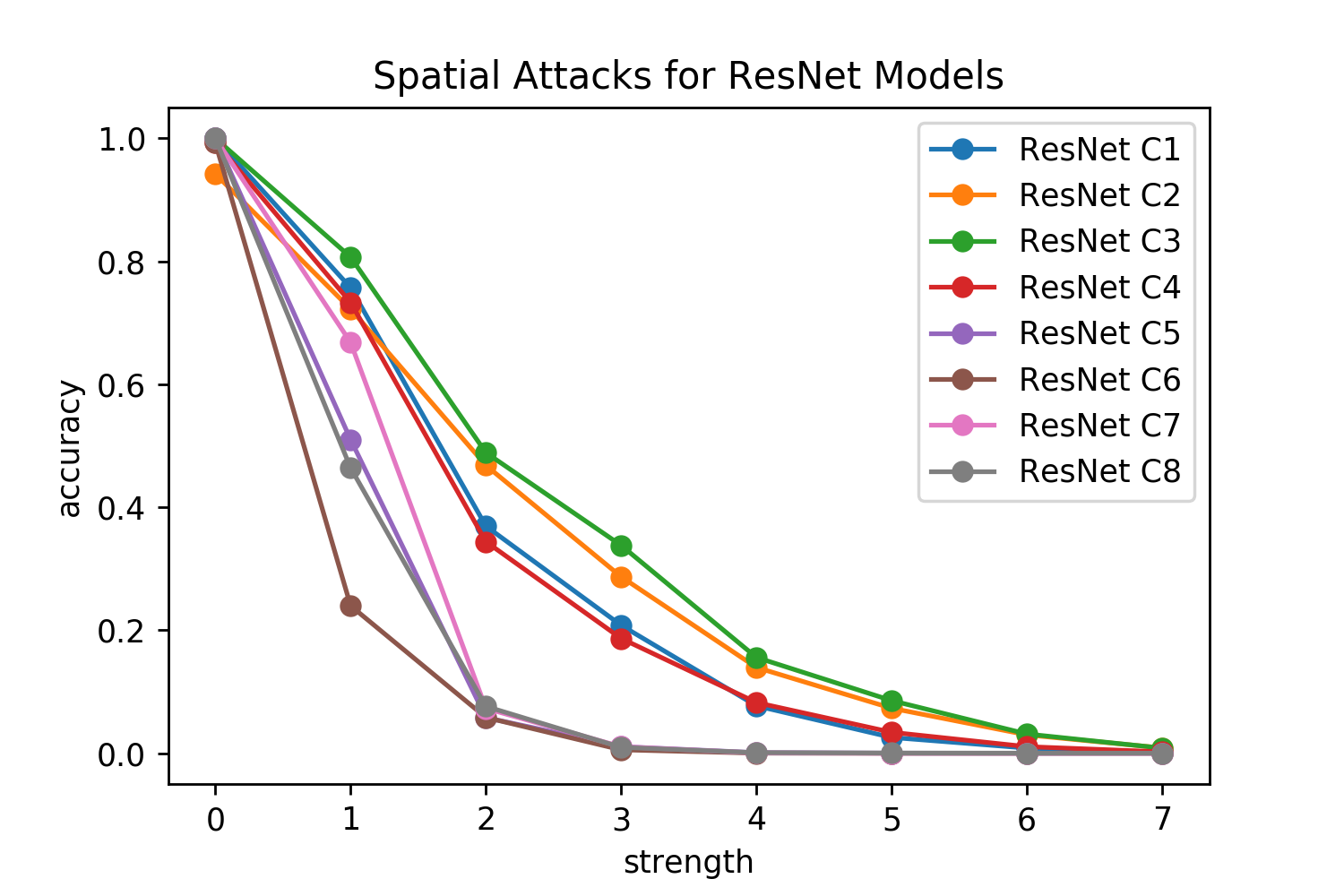}

		\caption{Accuracies of the retrained models for the proposed approaches.}
		\label{fig:pmv_results}
	\end{figure*}

	Figure \ref{fig:pmv_results} demonstrates the accuracies of retrained VGG and ResNet models for the proposed approaches.
	
	\textbf{PMV} led to an obvious separation between models with and without overfitting. The decrease rate of the models with the good fit (C1 and C3) differs significantly from the obviously overfitted models (C6 and C8). Interestingly, we observe three model clusters for ResNet architecture: models with good (C1, C3), medium (C2, C5), and bad (C4, C6-C8) fit.
	
	\textbf{PTV} resulted in similar accuracy drops as PMV. However, the ResNet models form two clusters: the models with the good fit are the same as with PMV (C1 and C3), whereas all further models look overfitted according to the accuracy drop.
	
	\textbf{FGSM Attack} led to a visible separation of models for ResNet. However, it failed to provide meaningful model differentiation for VGG models.
	
	\textbf{Gaussian Noise} was not that successful in differentiating models both for VGG and ResNet. Similarly to FGSM, the results for VGG were worse than those for ResNet models.
	
	\textbf{Spatial Attacks} turned out to be most informative: the behavior of models with a smaller gap between training and validation accuracy significantly differs from that of the models with a worse fit. Also, this is a single method that provides the evident separation of the VGG models.
	
\begin{figure}[t]
		\centering
		\includegraphics[width=0.9\linewidth]{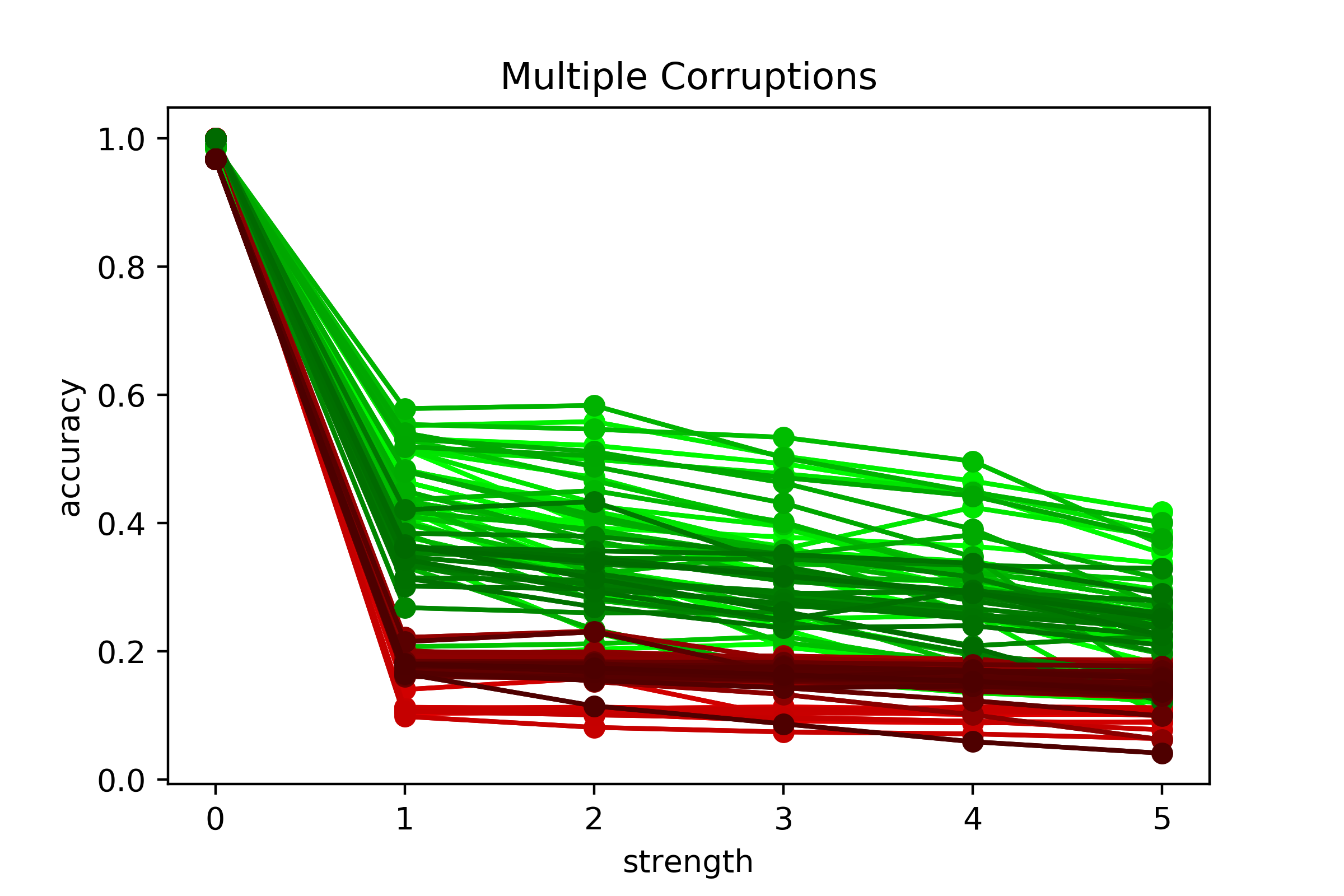}
		\caption{Accuracies of the retrained models for the Hedrycks corruptions approach. Models C1-C4 are plotted in green, models C5-C8 are plotted in red.}
		\label{fig:multiple_corruptions}
	\end{figure}

	\textbf{Hendrycks Corruptions} While the set of Hendrycks corruptions did behave differently for models with large and small accuracy gaps, the difference was not significant enough to confidently separate models based on their overfitting behavior. Also, no significant difference between various corruption types was found. Figure \ref{fig:multiple_corruptions} shows results combined for all corruption types.
	
	Overall, visual assessment of the graphs has revealed, that PMV, PTV and spatial attacks lead to the best separation of overfitted and correctly fitted models.

	\subsection{Overfitting Metrics}
	
	Tables \ref{tab:vgg_metrics} and \ref{tab:resnet_metrics} demonstrate the results for the studied overfitting metrics: (1) the linear fit slope $\hat{k}_S$ for PMV as introduced by Zhang et al.~\cite{zhang2019perturbed}, (2) $\hat{k}^{PTV}_S$, which is an updated version of it for PTV, (3) the proposed maximum decrease metric $\hat{k}_{max}(S,h)$, and (4) the proposed SSE metric $\widehat{SSE}$. The latter three metrics are reported for the Gaussian noise, FGSM and spatial attacks. 
	
	We could apply the $\hat{k}_{max}(S,h)$  and $\widehat{SSE}$ metrics to the multiple corruptions approach since there are no consecutive perturbation strengths with equal distanc. Instead, they are empirically determined by Hendrycks et al.~\cite{hendrycks2019benchmarking}. 
	
	Our experiments show, that apart from the Gaussian noise approach, all metrics consistently select models C1-C4 as the models with the best fit. . Moreover, we observe that in the case of ResNet models, each method apart from the Gaussian noise approach tends to select the same model both for maximum decrease and sum of squared error metrics. 
	
	Furthermore, PTV, as well as maximum decrease and SSE for spatial attacks and FGSM, consider C1 and C3 to be the VGG models with the best fit. For ResNet, different metrics select models C1-C4. Also, the worst (i.e. the most overfitted) models according to different metrics are C7 or C8 for VGG and C6 or C7 for ResNet models. However, no single model among C1-C4 is consistently selected as the best fit model by all approaches and metrics.

	If we consider model C1 to have the best fit, we conclude that the PTV and spatial attack approaches lead to the best results. Interestingly, this is only partially consistent with our visual assessment results, reported above.
	
	Overall, it seems that the models trained with more data (C1-C4) are generally less overfitted than those trained with less data (C5-C8).  Within a set of models trained with the same number of instances, the regularized models are less overfitted. Compared to the difference between training and validation accuracy, the proposed metrics offer a fine-grained assessment of the overfitting behavior of the examined models.
	
	\subsection{Evaluation on Models with Unknown Overfitting Behavior}
\begin{figure}[t]
		\centering
		\includegraphics[width=0.9\linewidth]{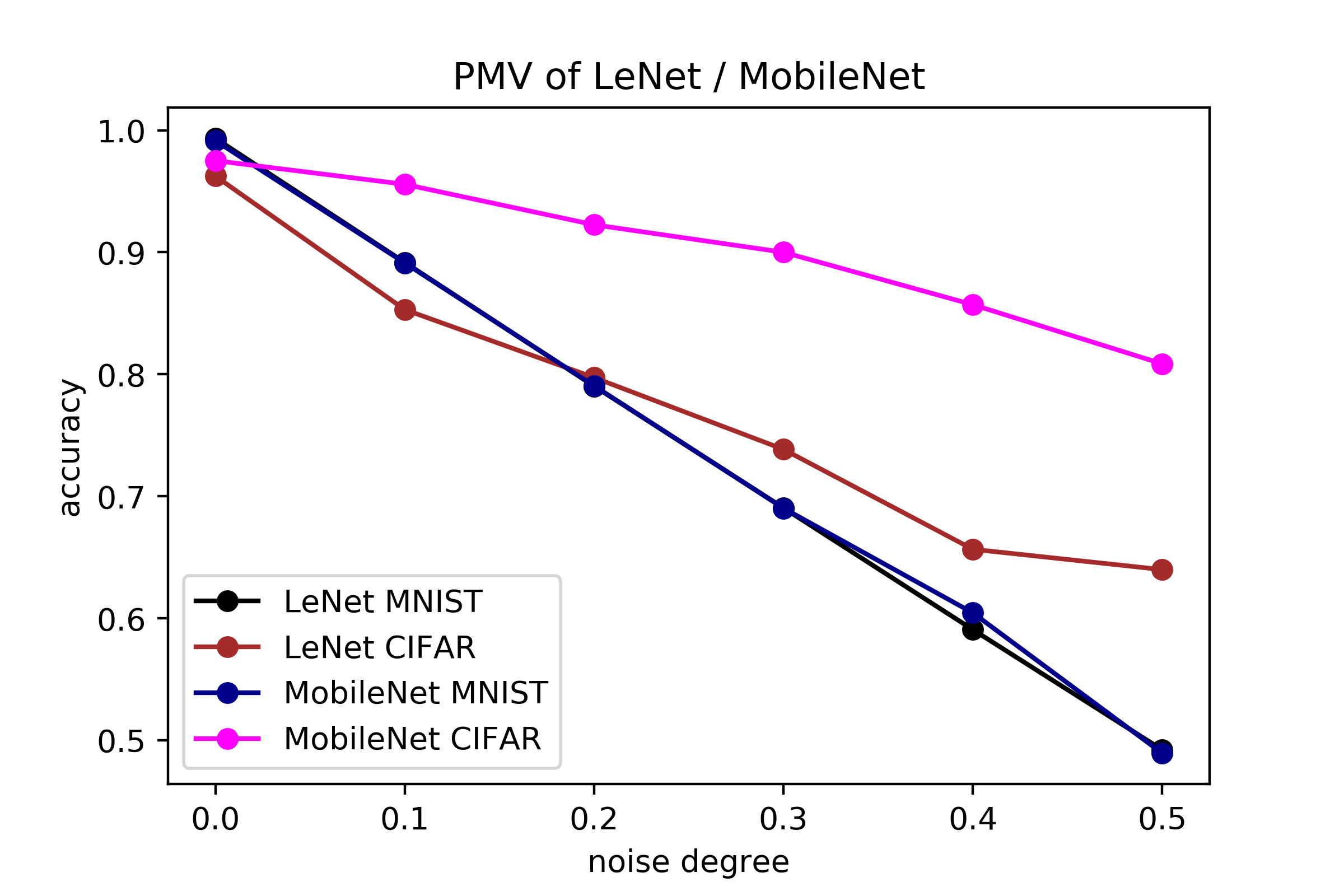}
		\includegraphics[width=0.9\linewidth]{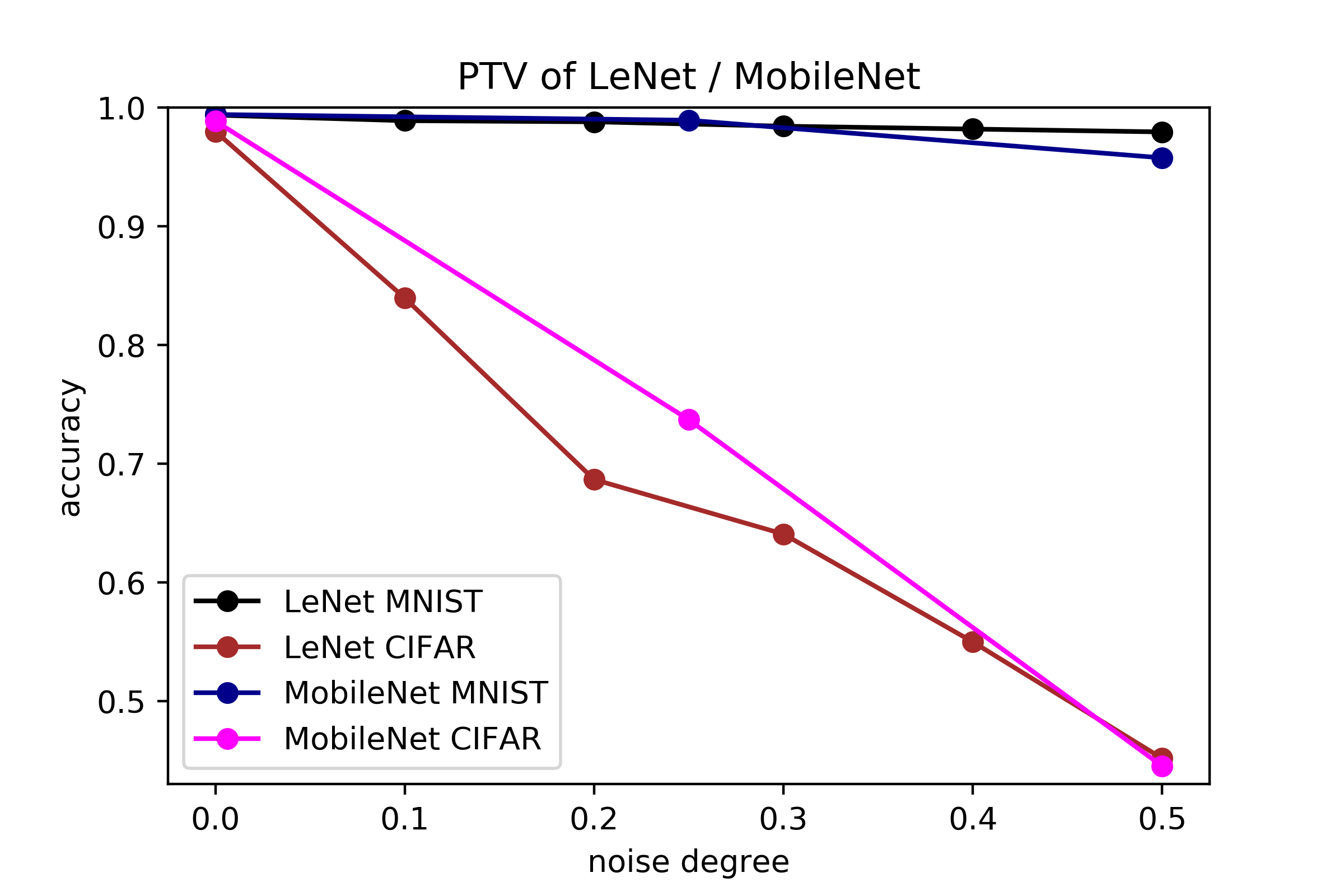}
		\includegraphics[width=0.9\linewidth]{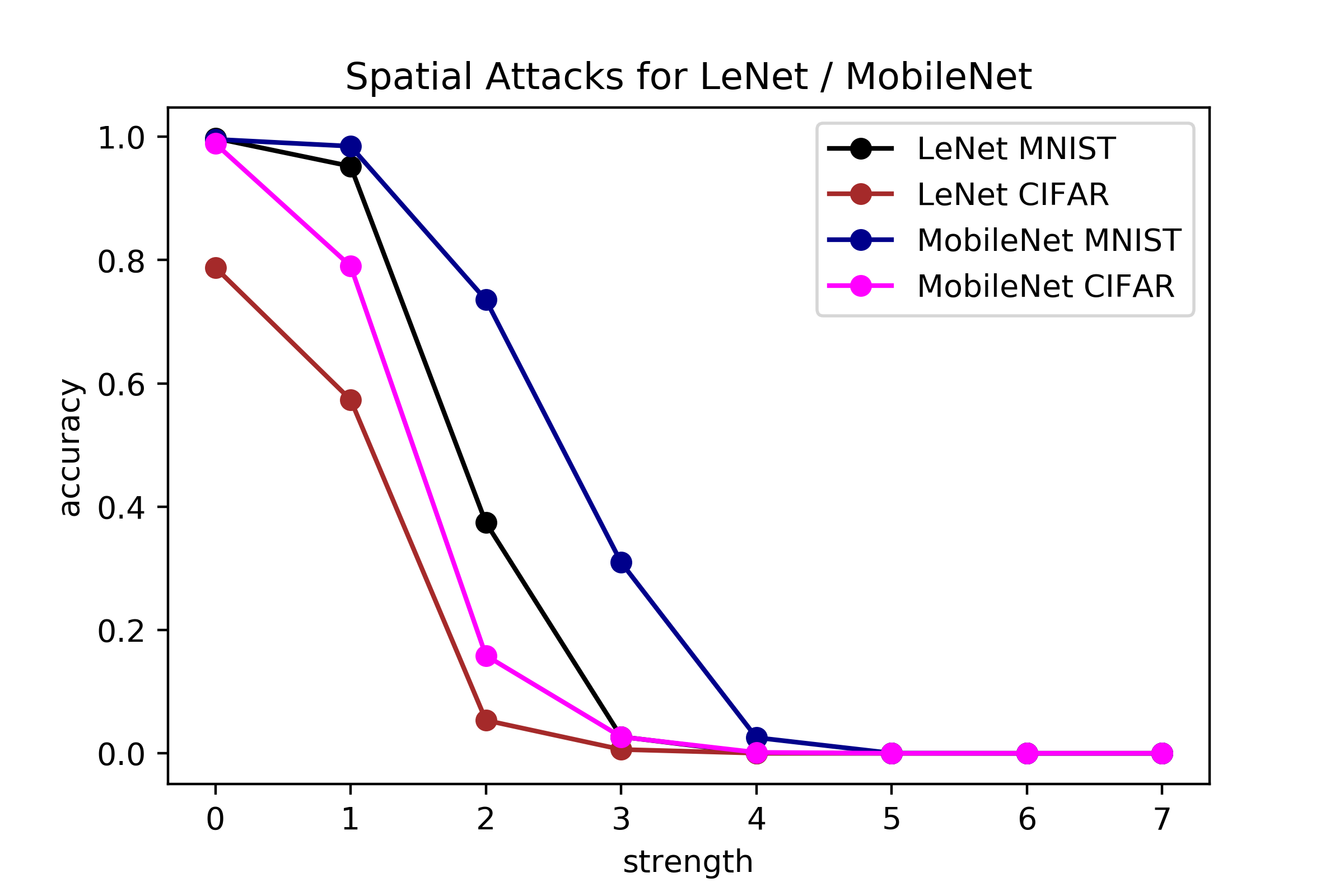}
		\caption{Accuracies of LeNet and MobileNet for the proposed approaches.}
		\label{fig:lenet_results}
	\end{figure}

	To assess the proposed metrics on further models beyond the defined pool of models, we run experiments with LeNet~\cite{lecun1998lenet} and MobileNet~\cite{howard2017mobilenets} trained on CIFAR-10 and MNIST~\cite{deng2012mnist}. No special changes have been introduced to the architectures or training procedures to enforce or avoid overfitting. We used standard Keras implementations for the experiments. Figure \ref{fig:lenet_results} shows accuracies of the LeNet and MobileNet models over different noise degrees for the proposed approaches.
	
	As seen in Table \ref{tab:lenet_results}, the metrics that most significantly separate the models are PMV and PTV. Both architectures seem to overfit against the CIFAR-10 dataset, but not against the MNIST dataset. It is further strengthened by the measured generalization gap. E.g. for LeNet, $\widehat{Acc}(S)- \widehat{Acc}(V)$ is large for CIFAR-10 ($0.4300$) and small for MNIST ($ 0.0062$). The spatial attack metrics do not differ as significantly, which prevents a distinction between the two models regarding the overfitting behavior.

	\begin{table}[t]
		\centering
		\caption{Results of LeNet and MobileNet models on MNIST and CIFAR-10 for different overfitting metrics}
		\label{tab:lenet_results}
		\begin{tabular}{|l|c|c|c|c|}
			\hline
			Approach			& PMV 	& PTV 		& Spatial  & Spatial    \\\hline
			Metric & & & $\hat{k}_{max}(S,h)$ & $\widehat{SSE}$ \\\hline
			LeNet MNIST		&	1.0022	&	0.0273	&0.5782							&0.3624									\\
			LeNet CIFAR		&	0.6458	&	1.0164	&0.5196							& 0.2651									\\ \hline
			MobileNet MNIST	&	0.9914	&	0.0730	&0.4260							&0.2034								\\
			MobileNet CIFAR	&0.3288		&	1.0870	&0.6318							& 0.3752									\\
			\hline
		\end{tabular}
	\end{table}
	
	Overall, the best separation of LeNet and MobileNet models was achieved with the PTV approach. For spatial attacks, the reported accuracies are slightly biased, due to the different accuracies at perturbation strength 0.

	\section{Conclusion}
	
	In this work, we have performed evaluation of CNNs for two existing overfitting detection approaches~\cite{werpachowski2019detecting} and~\cite{zhang2019perturbed}, that do not require a separate holdout set and rely on introducing label noise and adversarial examples to the training or test data. We have further extended these methods to include a broader spectrum of injected noise with the goal of measuring the overfitting to the training set of CNNs. Our pool of evaluated models comprised a total of 8 VGG and 8 ResNet models with various anti-overfitting features. 
	
	Based on the results of our experiments, we have introduced and evaluated new scalar overfitting metrics. They provide the possibility to compare and select models with different architectures and hyperparameter settings. Additionally, the metrics can be used to indicate whether a proposed neural network will perform as expected when confronted with unseen inputs. Two overfitting metrics can be retrieved without having the model code at hand, by just altering input data and observing the output. This facilitates the possibility to separate model development from testing, which might be crucial in sensitive scenarios. In particular, this makes it possible to detect overfitting of a model that is developed and provided by a third party and only accessible via API.
	
	Our evaluation has revealed, that the size of the training set has the largest impact on the overfitting behavior, followed by the regularization techniques such as weight regularization, batch normalization, and dropout. The number of parameters is, however, a less important factor in overfitting according to our findings. 
	
	Overall, the proposed metrics provide the fine-grained ranking of models according to their overfitting behavior, when compared to a standard technique of measuring the gap between training and validation accuracy. Additionally, it allows for comparison of models with different architectures and trained on different datasets. Maximum decrease and sum of squared error metrics, that have led to the most convincing results, in particular in combination with the spatial attack method, seem to be most promising for future research in this direction.
	
	\section*{Acknowledgment}
	
	This research was supported by the Ministry of Economic Affairs, Labour and Housing Baden-Württemberg and preformed within the project ”CyberProtect.”
	
	\bibliographystyle{IEEEtran}
	\bibliography{literature}

\begin{thebibliography}{10}
\providecommand{\url}[1]{#1}
\csname url@samestyle\endcsname
\providecommand{\newblock}{\relax}
\providecommand{\bibinfo}[2]{#2}
\providecommand{\BIBentrySTDinterwordspacing}{\spaceskip=0pt\relax}
\providecommand{\BIBentryALTinterwordstretchfactor}{4}
\providecommand{\BIBentryALTinterwordspacing}{\spaceskip=\fontdimen2\font plus
\BIBentryALTinterwordstretchfactor\fontdimen3\font minus
  \fontdimen4\font\relax}
\providecommand{\BIBforeignlanguage}[2]{{%
\expandafter\ifx\csname l@#1\endcsname\relax
\typeout{** WARNING: IEEEtran.bst: No hyphenation pattern has been}%
\typeout{** loaded for the language `#1'. Using the pattern for}%
\typeout{** the default language instead.}%
\else
\language=\csname l@#1\endcsname
\fi
#2}}
\providecommand{\BIBdecl}{\relax}
\BIBdecl

\bibitem{recht2018cifar}
B.~Recht, R.~Roelofs, L.~Schmidt, and V.~Shankar, ``Do cifar-10 classifiers
  generalize to cifar-10?'' \emph{CoRR}, vol. abs/1806.00451, 2018.

\bibitem{yeom2018privacy}
S.~Yeom, I.~Giacomelli, M.~Fredrikson, and S.~Jha, ``Privacy risk in machine
  learning: Analyzing the connection to overfitting,'' in \emph{Computer
  Security Foundations Symposium (CSF)}.\hskip 1em plus 0.5em minus 0.4em\relax
  IEEE, 2018.

\bibitem{zhang2019perturbed}
J.~M. Zhang, E.~T. Barr, B.~Guedj, M.~Harman, and J.~Shawe-Taylor, ``Perturbed
  model validation: A new framework to validate model relevance,'' \emph{CoRR},
  vol. abs/1905.10201, 2019.

\bibitem{werpachowski2019detecting}
R.~Werpachowski, A.~Gy{\"{o}}rgy, and C.~Szepesv{\'{a}}ri, ``Detecting
  overfitting via adversarial examples,'' in \emph{Advances in Neural
  Information Processing Systems (NIPS)}, 2019.

\bibitem{recht2019imagenet}
B.~Recht, R.~Roelofs, L.~Schmidt, and V.~Shankar, ``Do imagenet classifiers
  generalize to imagenet?'' in \emph{International Conference on Machine
  Learning (ICML)}, K.~Chaudhuri and R.~Salakhutdinov, Eds.\hskip 1em plus
  0.5em minus 0.4em\relax {PMLR}, 2019.

\bibitem{deng2012mnist}
L.~Deng, ``The mnist database of handwritten digit images for machine learning
  research [best of the web],'' \emph{IEEE Signal Processing Magazine}, 2012.

\bibitem{russakovsky2015imagenet}
O.~Russakovsky, J.~Deng, H.~Su, J.~Krause, S.~Satheesh, S.~Ma, Z.~Huang,
  A.~Karpathy, A.~Khosla, M.~S. Bernstein, A.~C. Berg, and L.~Fei{-}Fei,
  ``Imagenet large scale visual recognition challenge,'' \emph{International
  Journal of Computer Vision}, 2015.

\bibitem{simonyan2014vgg}
K.~Simonyan and A.~Zisserman, ``Very deep convolutional networks for
  large-scale image recognition,'' in \emph{International Conference on
  Learning Representations (ICLR)}, 2015.

\bibitem{krizhevsky2009cifar}
A.~Krizhevsky, G.~Hinton \emph{et~al.}, ``Learning multiple layers of features
  from tiny images,'' Citeseer, Tech. Rep., 2009.

\bibitem{he2016resnet}
K.~He, X.~Zhang, S.~Ren, and J.~Sun, ``Deep residual learning for image
  recognition,'' in \emph{Conference on Computer Vision and Pattern Recognition
  (CVPR)}.\hskip 1em plus 0.5em minus 0.4em\relax {IEEE} Computer Society,
  2016.

\bibitem{liu2015vggcifar}
S.~Liu and W.~Deng, ``Very deep convolutional neural network based image
  classification using small training sample size,'' in \emph{Asian Conference
  on Pattern Recognition (ACPR)}.\hskip 1em plus 0.5em minus 0.4em\relax IEEE,
  2015.

\bibitem{chollet2015keras}
F.~Chollet \emph{et~al.}, ``Keras,'' \url{https://keras.io}, 2015.

\bibitem{kingma2014adam}
D.~P. Kingma and J.~Ba, ``{Adam: A Method for Stochastic Optimization},''
  \emph{International Conference on Learning Representations (ICLR)}, 2015.

\bibitem{li2019dropoutBN}
X.~Li, S.~Chen, X.~Hu, and J.~Yang, ``Understanding the disharmony between
  dropout and batch normalization by variance shift,'' in \emph{Conference on
  Computer Vision and Pattern Recognition (CVPR)}.\hskip 1em plus 0.5em minus
  0.4em\relax Computer Vision Foundation / {IEEE}, 2019.

\bibitem{rolnick2017deeprobust}
D.~Rolnick, A.~Veit, S.~Belongie, and N.~Shavit, ``Deep learning is robust to
  massive label noise,'' \emph{CoRR}, vol. abs/1705.10694, 2017.

\bibitem{goodfellow2014explaining}
I.~J. Goodfellow, J.~Shlens, and C.~Szegedy, ``Explaining and harnessing
  adversarial examples,'' in \emph{International Conference on Learning
  Representations (ICLR)}, 2015.

\bibitem{madry2017towards}
A.~Madry, A.~Makelov, L.~Schmidt, D.~Tsipras, and A.~Vladu, ``{Towards Deep
  Learning Models Resistant to Adversarial Attacks},'' \emph{International
  Conference on Learning Representations (ICLR)}, 2018.

\bibitem{azulay2018smalltranslation}
A.~Azulay and Y.~Weiss, ``Why do deep convolutional networks generalize so
  poorly to small image transformations?'' \emph{J. Mach. Learn. Res.}, 2019.

\bibitem{engstrom2019exploring}
L.~Engstrom, B.~Tran, D.~Tsipras, L.~Schmidt, and A.~Madry, ``Exploring the
  landscape of spatial robustness,'' in \emph{International Conference on
  Machine Learning (ICML)}, 2019.

\bibitem{ford2019adversarialnoise}
J.~Gilmer, N.~Ford, N.~Carlini, and E.~D. Cubuk, ``Adversarial examples are a
  natural consequence of test error in noise,'' in \emph{International
  Conference on Machine Learning (ICML)}.\hskip 1em plus 0.5em minus
  0.4em\relax {PMLR}, 2019.

\bibitem{hendrycks2019benchmarking}
D.~Hendrycks and T.~G. Dietterich, ``Benchmarking neural network robustness to
  common corruptions and perturbations,'' in \emph{International Conference on
  Learning Representations (ICLR)}, 2019.

\bibitem{lecun1998lenet}
Y.~LeCun, L.~Bottou, Y.~Bengio, and P.~Haffner, ``Gradient-based learning
  applied to document recognition,'' \emph{Proceedings of the IEEE}, vol.~86,
  no.~11, pp. 2278--2324, 1998.

\bibitem{howard2017mobilenets}
A.~G. Howard, M.~Zhu, B.~Chen, D.~Kalenichenko, W.~Wang, T.~Weyand,
  M.~Andreetto, and H.~Adam, ``Mobilenets: Efficient convolutional neural
  networks for mobile vision applications,'' \emph{CoRR}, vol. abs/1704.04861,
  2017.

\end{thebibliography}

\end{document}